\documentclass{article}

\PassOptionsToPackage{numbers, compress}{natbib}

\usepackage[final]{neurips_2025}

\usepackage[utf8]{inputenc} 
\usepackage[T1]{fontenc}    
\usepackage{hyperref}       
\usepackage{url}  
\usepackage{graphicx}
\usepackage{booktabs}       
\usepackage{amsfonts}       
\usepackage{nicefrac}       
\usepackage{microtype}      
\usepackage{xcolor}         
\usepackage{xpatch}
\usepackage{amsmath}
\usepackage{float}
\usepackage{pifont}
\usepackage{multirow}
\usepackage{changes}

\makeatletter
\xapptocmd{\NAT@bibsetnum}{\setlength{\leftmargin}{0pt}\setlength{\itemindent}{\labelwidth}\addtolength{\itemindent}{\labelsep}}{}{}
\makeatother

\title{4DGCPro: Efficient Hierarchical 4D Gaussian Compression for Progressive Volumetric Video Streaming}

\author{%
  Zihan Zheng$^{1}$ , Zhenlong Wu$^{1}$ , Houqiang Zhong$^{2}$ , Yuan Tian$^{2,3}$ , Ning Cao$^{4}$ , \\ \textbf{Lan Xu$^{5}$ , Jiangchao Yao$^{1}$ , Xiaoyun Zhang$^{1}$ , Qiang Hu$^{1}$\thanks{The corresponding author is Qiang Hu(qiang.hu@sjtu.edu.cn)} , Wenjun Zhang$^{1,2}$} \\
  Cooperative Medianet Innovation Center, Shanghai Jiaotong University$^{1}$\\
  Department of Electronics, Shanghai Jiaotong University$^{2}$\\
  Shanghai AI Lab$^{3}$ \\
  Cloud platform department, E-surfing Vision Technology Co., Ltd.$^{4}$ \\
  School of Information Science and Technology, ShanghaiTech University$^{5}$ \\
}

\begin{document}

\setlength{\lineskiplimit}{0pt}
\setlength{\lineskip}{0pt}
\setlength{\abovedisplayskip}{3pt}  
\setlength{\belowdisplayskip}{3pt}
\setlength{\abovedisplayshortskip}{3pt}
\setlength{\belowdisplayshortskip}{3pt}

\maketitle
\begin{figure*}[h]
\centering
\vspace{-3em}
\includegraphics[width=\linewidth]{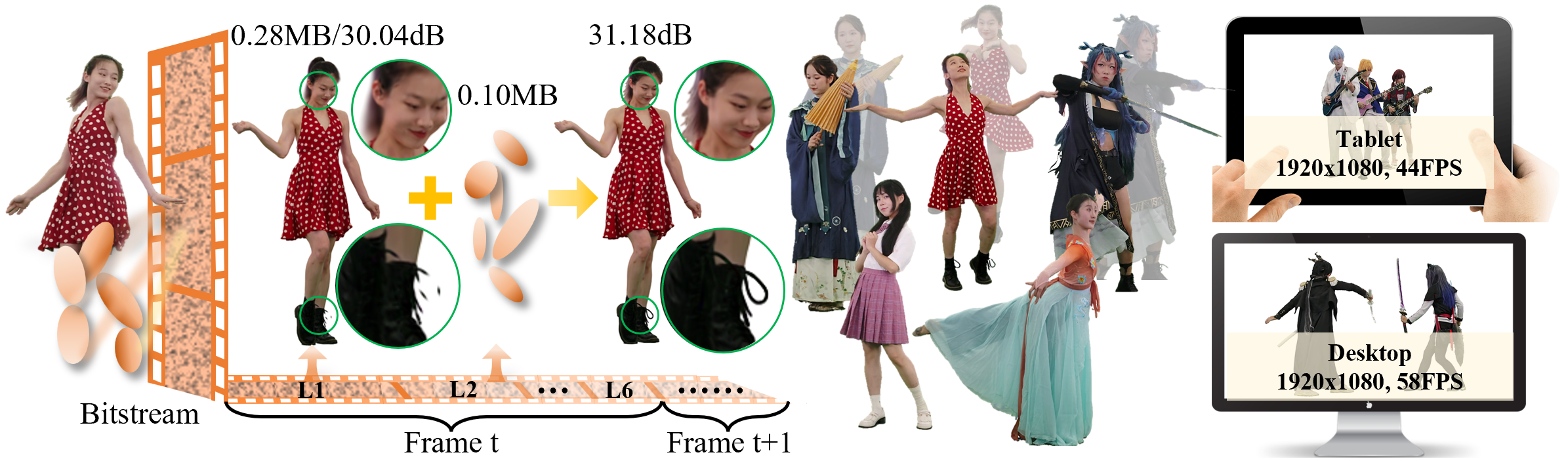}
\vspace{-1.5em}
\caption{\textbf{Left:} Our method enables progressive streaming of hierarchical 4D Gaussians within a single bitstream, where incremental enhancement layers (e.g., +0.10MB) gradually improve visual quality (e.g., from 30.04dB to 31.18dB) with minimal bitrate overhead. \textbf{Right:} The streamed content is adaptively decoded and rendered in real-time on various devices (e.g., tablets 44FPS, desktops 58FPS) by dynamically selecting layers (L1–L6) based on available bandwidth and compute.}
\vspace{-0.5em}
\label{fig:teaser}
\end{figure*}


\begin{abstract}
\vspace{-0.5em}
Achieving seamless viewing of high-fidelity volumetric video, comparable to 2D video experiences, remains an open challenge. Existing volumetric video compression methods either lack the flexibility to adjust quality and bitrate within a single model for efficient streaming across diverse networks and devices, or struggle with real-time decoding and rendering on lightweight mobile platforms. To address these challenges, we introduce 4DGCPro, a novel hierarchical 4D Gaussian compression framework that facilitates real-time mobile decoding and high-quality rendering via progressive volumetric video streaming in a single bitstream. Specifically, we propose a perceptually-weighted and compression-friendly hierarchical 4D Gaussian representation with motion-aware adaptive grouping to reduce temporal redundancy, preserve coherence, and enable scalable multi-level detail streaming. Furthermore, we present an end-to-end entropy-optimized training scheme, which incorporates layer-wise rate-distortion (RD) supervision and attribute-specific entropy modeling for efficient bitstream generation. Extensive experiments show that 4DGCPro enables flexible quality and multiple bitrate within a single model, achieving real-time decoding and rendering on mobile devices while outperforming existing methods in RD performance across multiple datasets. Project Page: \href{https://mediax-sjtu.github.io/4DGCPro}{https://mediax-sjtu.github.io/4DGCPro}
\vspace{-0.5em}
\end{abstract}


\section{Introduction}



Volumetric video enables immersive 3D experiences with free-viewpoint navigation, but streaming and rendering high-quality long sequences with large motions remains challenging, especially on lightweight devices like mobile phones. Compared to 2D video, volumetric content demands higher bandwidth, storage, and real-time decoding capabilities, making fixed-bitrate solutions inadequate to handle the variability across heterogeneous devices and network conditions. Therefore, the core challenge lies in achieving real-time, high-fidelity playback with low computational cost, while enabling scalable and progressive streaming under constrained resources.

Traditional volumetric reconstruction methods, including surface estimation \cite{dai2017bundlefusion}, point clouds \cite{schoenberger2016sfm}, meshes \cite{Wang_2018_ECCV}, light field \cite{LightFieldRendering, panoramiclightfieldsystem} and depth-based techniques \cite{RobustFusion,kinectsensor}, struggle to faithfully capture the geometric complexity and temporal dynamics of real-world scenes. Neural radiance field (NeRF) \cite{nerf} address these limitations by modeling view-dependent appearance without relying on explicit geometry, enabling photorealistic rendering.  While extensions \cite{DNeRF,streaming,tineuvox,HumanRF,kplanes,hexplane_2023_CVPR,zhang2021editable} adapt NeRF to dynamic scenes, they remain constrained by the difficulty of handling long sequences and efficient streaming. Some works \cite{rerf, videorf, tetrirf, zheng2024jointrf, zheng2024hpc} compress dynamic NeRFs to enable streaming, but the high computational cost of decoding and rendering limits their practicality in real-time applications.

Recent work on 3D Gaussian Splatting (3DGS) \cite{kerbl3Dgaussians} introduces an explicit scene representation using anisotropic Gaussian primitives with real-time, differentiable rasterization, achieving unprecedented rendering speed and visual quality. Subsequent studies \cite{Li_STG_2024_CVPR,Wu_2024_CVPR,Guo2024Motionaware3G,yan20244d} extend 3DGS to dynamic scenes by incorporating temporal attributes into Gaussian parameters, but require full-sequence pre-loading during training and rendering, limiting streaming practicality. Alternative approaches \cite{luiten2023dynamic,sun20243dgstream,NEURIPS2024_4c9477b9,hicom2024} model temporal Gaussian variations via deformable fields or residual tracking to enable streamable representations, yet incur substantial bandwidth overhead.  A few studies \cite{hu20254dgcrateaware4dgaussian,hifi4g, jiang2024robust,wang2024v} have explored compression for dynamic 3DGS. For example, 4DGC \cite{hu20254dgcrateaware4dgaussian} jointly optimizes representation and entropy models via RD loss, improving efficiency but struggling with high decoding latency and poor robustness to large motions due to rigid modeling. More fundamentally, existing dynamic Gaussian compression methods lack the flexibility to adjust video quality and bitrate within a single model, and typically require separate models for each bitrate, leading to high storage costs and limited adaptability under varying network and device conditions.

To tackle the above challenges, we propose 4DGCPro, a novel hierarchical 4D Gaussian compression approach for  progressive volumetric video streaming. As illustrated in Fig. \ref{fig:teaser}, our method achieves multiple bitrate using a single model and enables real-time decoding and high-fidelity rendering on lightweight devices for large-motion sequences. We realize this through three key innovations. First, we introduce a perceptually-weighted hierarchical Gaussian representation for keyframes, guided by a significance metric that combines geometric volume and opacity. This enables scalable representation across detail levels and establishes the foundation for dynamic modeling. Second, we propose a hierarchical motion modeling strategy, where motion in subsequent frames is decomposed into rigid transformations and residual deformations to capture large displacements and preserve temporal coherence. We further adopt motion-aware adaptive Gaussian grouping to handle topological changes and long-term dynamics, ensuring compact and consistent temporal representation. 

Third, we propose a joint entropy-optimized training and progressive coding framework for efficient and scalable bitstream generation. Specifically, we introduce layer-wise rate-distortion (RD) supervision into the training pipeline using differentiable quantization and attribute-specific entropy modeling. For keyframes, we utilize FFT-accelerated Gaussian kernel density estimation (KDE) for precise bitrate prediction of Gaussian attributes, with hierarchical Gaussian optimization guided by per-layer RD trade-offs. For inter-frame, we apply Gaussian-distribution-based entropy estimation and temporal consistency constraints to maintain compactness and coherence. After training, we quantize attributes and convert multi-layer representations into stacked 2D single-channel maps, which are encoded with 2D codecs into a progressive bitstream, enabling scalable real-time decoding and rendering via hardware video codecs and shaders. Experimental results show that our 4DGCPro supports multiple bitrates using a single model and achieves state-of-the-art RD performance across various datasets. Compared to the SOTA method HPC \cite{zheng2024hpc}, our approach achieves a \textbf{3x} compression rate without quality degradation, while enabling real-time decoding and rendering on mobile devices.

In summary, our contributions are as follows:

\begin{itemize}
    \item We propose 4DGCPro, a novel framework for progressive volumetric video streaming that supports multiple bitrates with a single compact model, enabling real-time decoding and rendering on mobile devices with superior RD performance.
    \item We introduce a compact  hierarchical 4D Gaussian representation with motion-aware adaptive grouping for scalable and high-fidelity modeling of dynamic scenes.
    \item We present an end-to-end entropy-optimized training scheme with layer-wise RD supervision and attribute-specific entropy modeling, enabling fine-grained RD optimization across layers and better overall compression.
\end{itemize}

\section{Related Work}
\subsection{NeRF-based Volumetric Video Modeling}




NeRF \cite{nerf} have revolutionized 3D scene representation using differentiable volume rendering with implicit neural representations. While recent advances in static scene representation \cite{barron2021mipnerf,barron2022mipnerf360,barron2023zipnerf,DBARF_Chen_2023_CVPR,martinbrualla2020nerfw,instant-ngp,park2023camp,fpo++,merf} have improved compactness and reconstruction speed, several works have extended these methods to dynamic scenes.
Flow-based approaches \cite{9578364,Li_2023_CVPR} construct 3D features from monocular video, reducing data collection complexity but requiring additional priors 
for complex scenes. Deformation field methods \cite{NeuralRadianceFlow,Nerfies,DNeRF,nerfplayer} warp dynamic frames into a canonical space to capture temporal features, yet suffer from slow training and rendering. To accelerate performance, recent methods \cite{tineuvox,HumanRF,kplanes,hexplane_2023_CVPR,tensor4d,li2022neural,Park2023TemporalII,10377294,9879176} adopt explicit 4D radiance field representations based on structured volumetric decompositions (e.g., voxel grids, multi-plane projections, or tensor factorizations), yet these unified frameworks remain incompatible with streaming scenarios. 

\subsection{3DGS-based Volumetric Video Modeling}

3DGS \cite{kerbl3Dgaussians} and its variants \cite{Huang2DGS2024, feng2024flashgsefficient3dgaussian, höllein20243dgslmfastergaussiansplattingoptimization, charatan23pixelsplat, gao20246dgsenhanceddirectionawaregaussian}enable photorealistic static scene reconstruction through their efficiency and physical interpretability, with recent extensions to dynamic scenes. Current dynamic 3DGS approaches mainly follow two paradigms. Some studies \cite{Li_STG_2024_CVPR,Wu_2024_CVPR,yang2023deformable3dgs,Guo2024Motionaware3G,yan20244d,xu20244k4d,yang2023gs4d} model Gaussian attributes as temporal functions to create unified dynamic representations, which achieve exceptional RD performance but neglect streaming feasibility. Alternative approaches \cite{luiten2023dynamic,sun20243dgstream,NEURIPS2024_4c9477b9,hicom2024} employ frame-wise modeling with explicit rigid motion estimation, enabling streamable Gaussian volumetric video at the cost of increased data volume and compromised reconstruction quality. While $\text{V}^{3}$ \cite{wang2024v} optimizes the full Gaussian coefficients to model complex motions, its fixed group length leads to error accumulation or data redundancy. Meanwhile, it lacks the capability to support multiple-bitrate selection within a single bitstream.
Our approach introduces a compact hierarchical motion-aware Gaussian representation coupled with adaptive Gaussian grouping that dynamically responds to topological changes and achieves multiple bitrate using a single model.

\subsection{Volumetric Video Compression}

Volumetric video compression is crucial for reducing massive data requirements, where traditional approaches employ octree \cite{10.5555/2386388.2386404,7405340} and wavelet \cite{Wavelet-based} techniques (later standardized as MPEG-PCC \cite{8571288}), while subsequent learning-based methods \cite{10.1145/3512527.3531423,8803413,9287077,8786879,9459517,OctAttention,9321375,10960681} focus on improved efficiency. While recent advances \cite{rerf,videorf,tetrirf,nerfplayer,peng2023representing,cut,miniwave,zheng2024jointrf,zheng2024hpc,vrvvc} have made progress in compressing dynamic NeRF features for storage optimization, they commonly suffer from poor quality and slow decoding/rendering. For instance, HPC \cite{zheng2024hpc} employs learned compression for progressive coding of residual feature grids representing dynamic scenes. However, its high decoding latency limits real-time applications. 
For 3DGS-based methods, static scene techniques \cite{navaneet2023compact3d,fan2023lightgaussian,scaffoldgs,wang2024contextgs} dominate, whereas dynamic scene approaches \cite{hifi4g,jiang2024robust,hu20254dgcrateaware4dgaussian} face inefficiency and single-rate constraints. 
Our method delivers superior RD performance and computational efficiency using standard video codecs, supporting both hardware-accelerated real-time decoding and progressive streaming for quality adaptation across dynamic bandwidth.

\section{Method}
\begin{figure*}[t]
\centering
\includegraphics[width=\linewidth]{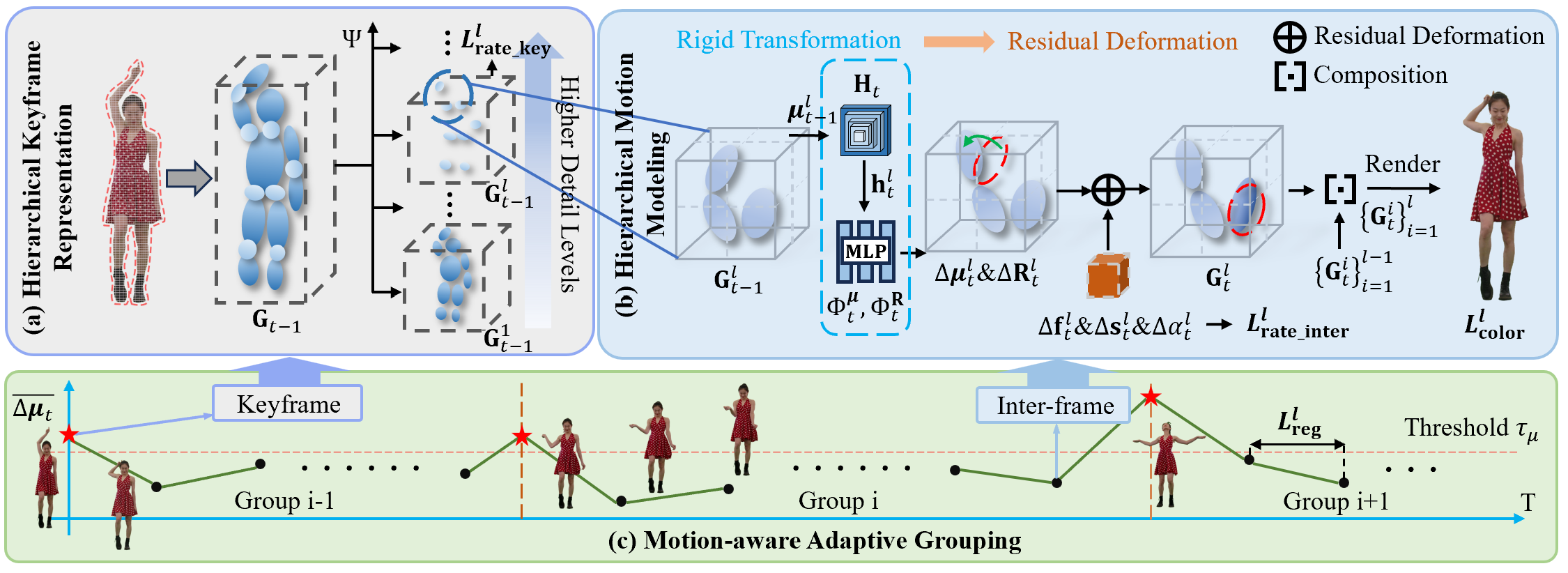}
\vspace{-2em}
\caption{\textbf{Our 4DGCPro framework.} (\textbf{a}) Perceptually-weighted hierarchical 4D Gaussian representation models keyframes at multi-level detail for progressive reconstruction. 
(\textbf{b}) Hierarchical motion modeling decomposes dynamic scenes into rigid transformations and residual deformations based on the previous frame, while (\textbf{c}) motion-aware  adaptive grouping dynamically adjusts to topological changes to enhance temporal consistency and reduce error accumulation. The entire pipeline is end-to-end optimized with layer-wise RD supervision and attribute-specific entropy modeling.}
\vspace{-1.8em}
\label{fig:method}
\end{figure*}


In this section, we present the technical details of our 4DGCPro architecture (Fig. \ref{fig:method}). The framework begins with a perceptually-weighted hierarchical Gaussian representation for keyframes (Sec. \ref{sec:3.1}), which establishes the foundation for dynamic scene characterization and progressive streaming. We then introduce a hierarchical motion modeling approach with adaptive grouping (Sec. \ref{sec:3.2}), decomposing motions into rigid transformations and residual deformations. This motion-aware adaptive Gaussian grouping mechanism effectively handles diverse motion patterns in complex scenes. To generate efficient and scalable bitstreams, we incorporate layer-wise RD optimization into the training pipeline through differentiable quantization and attribute-specific entropy modeling, followed by compression using standard 2D video codecs (Sec. \ref{sec:3.3}).
\vspace{-0.5em}

\subsection{Perceptually-Weighted Hierarchical Gaussian Keyframe Representation}
\label{sec:3.1}
Recall that 3DGS represents scenes explicitly through 3D Gaussians
$\mathbf{G}$, defined by a set of learnable parameters, including center position $\boldsymbol{\mu}$, rotation matrix $\mathbf{R}$ representing orientation, spherical harmonic coefficients $\mathbf{f}$ for view-dependent appearance modeling, scaling factors $\mathbf{s}$ controlling spatial extent, and opacity value $\alpha$. 
The spatial influence at point $\mathbf{x}$ follows $\mathbf {G}(\mathbf{x})$, expressed as:
\begin{align}
\setlength{\abovedisplayskip}{3pt}
\setlength{\belowdisplayskip}{3pt}
\mathbf{G}(\mathbf{x}) = \exp\left(-\frac{1}{2}(\mathbf{x}- \boldsymbol{\mu})^T\mathbf{\Sigma}^{-1}(\mathbf{x}-\boldsymbol{\mu})\right).  \label{equ_3dg_distribution}  
\end{align}
The covariance matrix $\mathbf{\Sigma} $ is constructed through $\mathbf{\Sigma} = \mathbf{R}\mathbf{s}\mathbf{s}^T\mathbf{R}^T$. With $\alpha_{i}'$ being the projection of the opacity of the $i$-th Gaussian onto the image plane and $\mathbf{c}_i$ denoting the color of the $i$-th Gaussian in the viewing direction, the pixel color $\mathbf{c}$ is computed by differentiable splatting of $N$ ordered Gaussians as follows:
\begin{align}
    \mathbf{c} = \sum_{i \in N} \mathbf{c}_i {\alpha'}_{i} \prod_{j=1}^{i-1} (1 - {\alpha'}_j),
    \label{render}
\end{align}
When reconstructing long-duration dynamic scenes, we first reconstruct high-quality keyframes to serve as references for subsequent inter-frames. Inspired by $\text{V}^{3}$ \cite{wang2024v}, we initialize keyframe 3D Gaussians through NeuS2-based \cite{neus2} surface mesh extraction. After pre-training, low-opacity Gaussians are pruned to achieve compact representations. While this optimized 3DGS delivers high-fidelity reconstruction, its substantial data footprint becomes problematic for smooth viewing under fluctuating bandwidth conditions. We therefore propose a perceptually-weighted hierarchical Gaussian representation guided by significance metric $\Psi$, which  serves as the basis for progressive transmission and rendering.
The proposed metric $\Psi$ analytically evaluates each Gaussian's visual importance through two geometrically-grounded attributes: (1) spatial volume $S$, representing the 3D volume occupied by the Gaussian and computed as $\frac{4}{3}\pi abc$ (where $a$, $b$, $c$ are its scale parameters along the three principal axes), which reflects its structural contribution to the scene geometry; and (2) opacity $\alpha$, which determines its perceptual weight in final rendering. These orthogonal factors are integrated with the weight $\lambda_{\Psi}$:
\begin{align}
\Psi = \alpha + \lambda_{\Psi} S.
\end{align}
After sorting all Gaussians in descending order using our significance metric, we partition them into $L$ hierarchical layers $\mathbf{G} = \{\mathbf{G}^l\}_{l=1}^L$. The base layer $\mathbf{G}^1$ preserves essential scene structures, while subsequent layers progressively enhance details. 
This hierarchical representation facilitates adaptive streaming, where the client dynamically selects the optimal number of layers $l$ to decode based on network conditions and computational resources.
This approach ensures smooth playback while efficiently balancing transmission overhead and reconstruction fidelity across diverse network environments.

\textbf{Progressive Rendering.} Our method supports progressive rendering from a single compressed representation, enabling scalable visual output with adjustable levels of detail. Starting from the base layer ($l=1$), which contains essential structural and appearance information, each subsequent Gaussian layer incrementally refines the reconstruction. Specifically, when decoding up to level $l$, only the Gaussians up to that layer (denoted by the index set $N^l$) are used for rendering. The color $\mathbf{c}^l$ at this stage is computed as:

\begin{align}
\setlength{\abovedisplayskip}{3pt}
\setlength{\belowdisplayskip}{3pt}
\mathbf{c}^l = \sum_{i \in N^l} \mathbf{c}^i {\alpha'}_{i} \prod_{j=1}^{i-1} (1 - {\alpha'}_j).
\label{progressive_render}
\end{align}

Each additional layer introduces a compact set of Gaussians that enhance detail without redundant transmission. This hierarchical refinement strategy allows the model to adapt dynamically to available computational and bandwidth resources, balancing reconstruction quality and efficiency in real time. The result is a scalable, high-fidelity rendering system capable of maintaining seamless visual enhancement under varying resource constraints, all within a unified bitstream.

\subsection{Hierarchical Motion Modeling with Adaptive Grouping}
\label{sec:3.2}
\begin{figure*}[t]
\centering
\includegraphics[width=\linewidth]{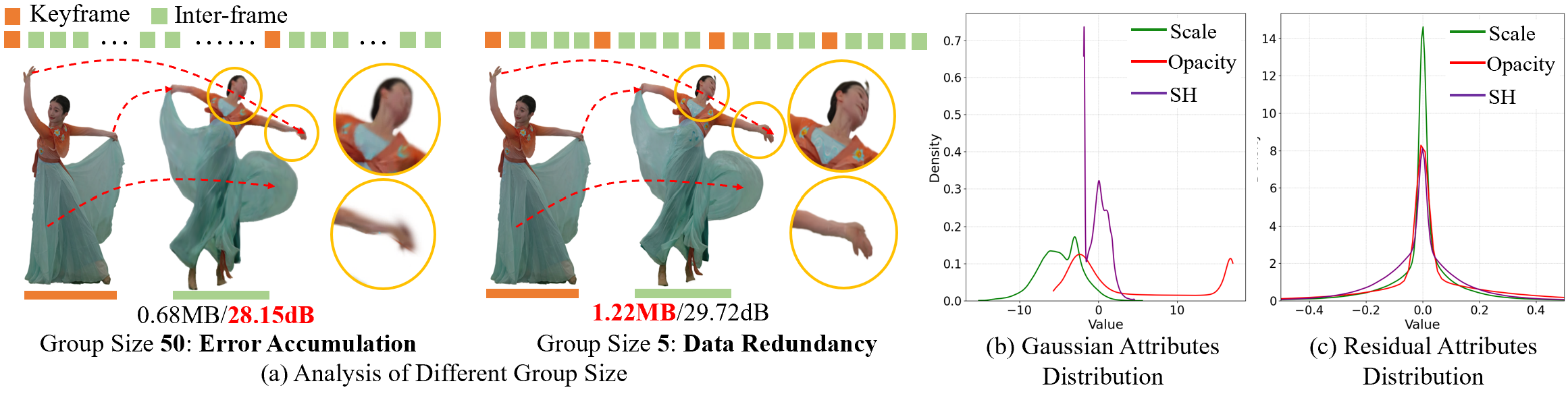}
\vspace{-2em}
\caption{Analysis of group size and attributes distributions.  \textbf{(a)} Large groups suffer from error accumulation while small groups exhibit data redundancy. \textbf{(b)} Keyframe Gaussian attributes display irregular spatial distributions, whereas \textbf{(c)} residual attributes follow Gaussian distributions.}
\label{fig:motivation}
\vspace{-1.5em}
\end{figure*}
Building upon the hierarchical keyframe Gaussian representation, we employ it as a reference basis for training subsequent inter-frames. Our proposed hierarchical motion modeling strategy effectively captures large-scale complex motions while maintaining temporal coherence through decomposition of frame differences into rigid transformations and residual deformations. 
We futher introduces motion-aware adaptive Gaussian grouping, which dynamically responds to varying scene changes to achieve dual benefits: enhanced representation fidelity and reduced model size for efficient streaming.

\textbf{Rigid Transformation.} To estimate the rigid transformations of Gaussians between frames, we utilize the Gaussian positions $\boldsymbol{\mu}_{t-1}=\{\boldsymbol{\mu}_{t-1}^l\}_{l=1}^L$ from the previous frame as input and predicts both translation $\Delta \boldsymbol{\mu}_{t}=\{\Delta \boldsymbol{\mu}_{t}^l\}_{l=1}^L$ and rotation $\Delta \mathbf{R}_t=\{\Delta \mathbf{R}_t^l\}_{l=1}^L$. The module first employs a multi-resolution hash grid $\mathbf{H}_t = \{ \mathbf{H}_t^l \}_{l=1}^{L_h}$ with $L_h$ levels to capture motion features $\mathbf{h}_t$ at different scales through hash coding as:
\begin{align}
\setlength{\abovedisplayskip}{3pt}
\setlength{\belowdisplayskip}{3pt}
    \mathbf{h}_t^l =  \{ \mathbf{h}_t^{l_h} \}_{l_h=1}^{L_h} = \{\text{interp}(\boldsymbol{\mu}_{t-1}^l,\mathbf{H}_t^{l_h}) \}_{{l_h}=1}^{L_h},
    \label{hash_coding}
\end{align}
where interp(·) refers to the hash grid interpolation operation. Subsequently, $\mathbf{h}_{t}$ is input into two lightweight MLPs, namely $\Phi_t^{\boldsymbol{\mu}}$ and $\Phi_t^{\mathbf{R}}$, to calculate the translation $\Delta \boldsymbol{\mu}_t^l$ and rotation $\Delta \mathbf{R}_t^l$ for each Gaussian:
\begin{align}
\setlength{\abovedisplayskip}{3pt}
\setlength{\belowdisplayskip}{3pt}
    \Delta \boldsymbol{\mu}_{t}^l = \Phi_t^{\boldsymbol{\mu}}(\mathbf{h}_t^l), \quad \Delta \mathbf{R}_t^l = \Phi_t^{\mathbf{R}}(\mathbf{h}^l_t).
    \label{ME}
\end{align}
In this manner, the position and rotation of frame $t$ can be determined using the equations $\boldsymbol{\mu}_t^l = \boldsymbol{\mu}_{t - 1}^l+\Delta \boldsymbol{\mu}_t^l$ and $\mathbf{R}_t^l=\Delta \mathbf{R}_t^l \mathbf{R}_{t - 1}^l$. 

\textbf{Residual Deformation.} Existing motion-aware 3DGS streaming methods \cite{hu20254dgcrateaware4dgaussian,sun20243dgstream} primarily focus on rigid motion simulation and Gaussian compensation, which often fail to accommodate object deformation and frequently introduces visual artifacts and temporal instability. To address these limitations, our approach incorporates a residual deformation framework following rigid transformation. 
The framework learns Gaussian deformations via adaptive scaling, opacity and color adjustments while predicting attribute residuals ($\Delta \mathbf{s}_{t}^l,\Delta \alpha_{t}^l,\Delta \mathbf{f}_{t}^l$) relative to parameters of t-1 frame, ensuring both local detail preservation and temporal stability.
\begin{align}
\setlength{\abovedisplayskip}{3pt}
\setlength{\belowdisplayskip}{3pt}
     \mathbf{s}_{t}^l = \mathbf{s}_{t-1}^l + \Delta \mathbf{s}_{t}^l, \quad \alpha_{t}^l = \alpha_{t-1}^l + \Delta \alpha_{t}^l, \quad \mathbf{f}_{t}^l = \mathbf{f}_{t-1}^l + \Delta \mathbf{f}_{t}^l.
    \label{Gaussian Deformation}
\end{align}
By combining both rigid transformations and residual deformations, our method effectively captures both large displacements and subtle scene variations, significantly reducing visual artifacts while maintaining temporal coherence.

\textbf{Motion-aware Adaptive Gaussian Grouping.} For long-sequence dynamic scenes with substantial motion, using only the initial frame as reference becomes inadequate due to accumulating scene variations. Meanwhile, as shown in Fig. \ref{fig:motivation}(a), fixed-length group structures inevitably introduce two competing artifacts: error accumulation across frames in large groups, and data redundancy in small groups due to repeated parameter transmission. We address this through motion-aware adaptive Gaussian grouping, where the group size is dynamically determined by rigid transformation results. When the average Gaussian translation $\overline{\Delta \boldsymbol{\mu}_{t}}$ exceeds a predefined threshold $\tau_{\mu}$, indicating substantial scene changes, we initiate a new group with an updated reference frame. This adaptive grouping strategy automatically adjusts to motion intensity, employing shorter groups during rapid changes for better reference quality, while maintaining longer groups for stable segments to optimize compression efficiency. The resulting representation achieves both accuracy and compactness by balancing temporal coherence with adaptive topology updates.

In summary, our 4DGCPro dynamically structures the scene into variable-length groups for efficient temporal modeling. For a group starting at frame T with length N, we sequentially represent it as $\mathbf{G}_T,\{\Delta \boldsymbol{\mu}_t, \Delta \mathbf{R}_t, \Delta \mathbf{f}_{t},\Delta \mathbf{s}_{t},\Delta \alpha_{t}\}^{T+N-1}_{t=T+1} $, where $\mathbf{G}_T$ is the keyframe Gaussian and $\{\Delta \boldsymbol{\mu}_t, \Delta \mathbf{R}_t, \Delta \mathbf{f}_{t},\Delta \mathbf{s}_{t},\Delta \alpha_{t}\}$ are the hierarchical residual attributes. This design optimally exploits inter-frame similarities while preserving reconstruction quality under complex motions.

\subsection{End-to-end Entropy-optimized Training}
\label{sec:3.3}
We propose an end-to-end entropy-optimized training scheme, which attains the optimal RD performance by incorporating layer-wise RD supervision and attribute-specific entropy modeling. To facilitate gradient back-propagation, we utilize differentiable quantization along with attribute-specific entropy modeling method to accurately estimate the bitrates of diverse attributes. Furthermore, we carry out progressive compression with 2D codecs on the hierarchical representation of Gaussians, enabling scalable real-time decoding and rendering. Next, we will introduce keyframe optimization, inter-keyframe optimization, and progressive bitstream generation in details. 

\textbf{Keyframe Optimization.} During the optimization process of keyframes, we first use $\mathcal{L}_{color}$ as a supervision term to pretrain the Gaussians:
\begin{equation}
\setlength{\abovedisplayskip}{3pt}
\setlength{\belowdisplayskip}{3pt}
    \mathcal{L}_{\text{color}} = (1 - \lambda_{\text{ssim}})  \|\mathbf{c}_g  - \hat{\mathbf{c}}\|_1 + \lambda_{\text{ssim}}\mathcal{L}_{\text{D-SSIM}},
\end{equation}
where $\mathbf{c}_g$ and $\hat{\mathbf{c}}$ denote the ground truth and reconstructed colors respectively, and $ \lambda_{\text{ssim}}$ weights the D-SSIM\cite{dssim} metric. After pre-training and pruning, we hierarchically organize Gaussians and perform joint entropy-optimized hierarchical training to maximize the RD performance per level. To ensure differentiable gradient flow and enhance quantization robustness, we implement uniform noise injection $u \sim U\left(-\frac{1}{2q}, \frac{1}{2q}\right)$ to simulate quantization effects with step size $q$. Additionally, we introduce entropy estimation of Gaussian attributes into our loss function to improve compression efficiency. As shown in Fig. \ref{fig:motivation}(b), keyframe Gaussian attributes exhibit irregular spatial distributions, necessitating KDE-based density estimation. Our implementation first computes the cumulative distribution function (CDF) through Silverman-rule bandwidth selection and FFT convolution, then obtains the probability density function (PDF) via numerical differentiation of the CDF:
\begin{equation}
\setlength{\abovedisplayskip}{3pt}
\setlength{\belowdisplayskip}{3pt}
P_{\text{PMF}}(\hat{y}) = P_{\text{CDF}}(\hat{y} + \frac{1}{2}) - P_{\text{CDF}}(\hat{y} - \frac{1}{2}). \label{PMF}
\end{equation}
To ensure optimal RD performance at each level, the keyframe optimization loss function $\mathcal{L}_{\text{key}}$ is formulated as a weighted sum of per-level losses $\mathcal{L}^l_{\text{key}}$, where each level's loss combines a photometric term $\mathcal{L}^l_{\text{color}}$ and a rate term $\mathcal{L}_{\text{rate\_key}}^l$:
\begin{align}
    & \mathcal{L}_{\text{key}} = \sum_{l=1}^{L} \lambda^l \mathcal{L}^l_{\text{key}} = \sum_{l=1}^{L} \lambda^l (\mathcal{L}_{\text{color}}^l + \lambda_{\text{rate\_key}} \mathcal{L}^l_{\text{rate\_key}}),\\
    & \mathcal{L}^l_{\text{color}} = (1 - \lambda_{\text{ssim}})  \|\mathbf{c}_g  - \hat{\mathbf{c}}^l\|_1 + \lambda_{\text{ssim}}\mathcal{L}^l_{\text{D-SSIM}},\\
    & \mathcal{L}_{\text{rate\_key}}^l = -\frac{1}{N}\sum_{\hat{y}_t^l \in \{\hat{\mathbf{R}}^l_t,\hat{\mathbf{s}}^l_t,\hat{\mathbf{f}}^l_t,\hat{\alpha}^l_t\}}{\log_2\left(P_{\text{PMF}}(\hat{y}^l_t)\right)}.
\end{align}
Here, $\mathcal{L}^l_{\text{color}}$ measures the photometric difference between the ground truth and the Gaussian rendering results at level $l$, $\mathcal{L}_{\text{rate\_key}}^l$ denotes the entropy estimated via KDE from Gaussian attributes at the same level. $\lambda_{\text{rate\_key}}$ is the weight of the entropy loss, and $\lambda^l$ is the weight parameter for the loss at different levels.  With this training objective, we obtain the keyframe Gaussians that achieve the optimal RD performance at each level. 

\textbf{Inter-frame Optimization.} Building upon the hierarchically trained Gaussians of keyframes, we optimize subsequent Gaussians within each group. Since Gaussian positions and rotations are particularly crucial for rendering quality, we only employ simulated quantization deliberately exclude entropy constraints and rely solely on 
$\mathcal{L}_{\text{color}}$ for supervision.

In the residual deformation stage, to maximize both accuracy and compactness at each level, we maintain hierarchical supervision by augmenting color constraints with both entropy loss $\mathcal{L}^l_{\text{rate\_inter}}$ and temporal loss $\mathcal{L}^l_{\text{reg}}$. As illustrated in Fig. \ref{fig:motivation}(c), we validate the Gaussian distribution of residual attributes, which allows us to simplify the entropy estimation to merely calculating the mean and variance of residuals, significantly streamlining training. To further enhance temporal coherence, we impose the temporal loss on residual attributes, explicitly enforcing inter-frame consistency. This deliberate smoothness constraint not only improves reconstruction quality but also reduces residual magnitudes during subsequent coding, ultimately optimizing storage efficiency. Thus, the training objective $\mathcal{L}_{\text{inter}}$ for this stage can be summarized as:
\begin{align}
    & \mathcal{L}_{\text{inter}} = \sum_{l=1}^{L} \lambda^l \mathcal{L}^l_{\text{inter}} = \sum_{l=1}^{L} \lambda^l (\mathcal{L}_{\text{color}}^l + \lambda_{\text{rate\_inter}} \mathcal{L}^l_{\text{rate\_inter}} + \lambda_{\text{reg}} \mathcal{L}^l_{\text{reg}}),\\
    & \mathcal{L}_{\text{rate\_inter}}^l = -\frac{1}{N}\sum_{\hat{y}_t^l \in \{\Delta\hat{\mathbf{s}}^l_t,\Delta\hat{\mathbf{f}}^l_t,\Delta\hat{\alpha}^l_t\}}{\log_2\left(P_{\text{PMF}}(\hat{y}_t^l)\right)},\\
    & \mathcal{L}^l_{\text{reg}} = \sum_{\hat{y}_t^l \in \{\Delta\hat{\mathbf{s}}_t^l,\Delta\hat{\mathbf{f}}_t^l,\Delta\hat{\alpha}_t^l\}}||\hat{y}_t^l||_1,
\end{align}
where $\mathcal{L}^l_{\text{inter}}$ denotes the inter-frame loss for the $l$-th layer of Gaussians, while $\mathcal{L}^l_{\text{rate\_inter}}$ and $\mathcal{L}^l_{\text{reg}}$ are weighted by $\lambda_{\text{rate\_inter}}$ and $\lambda_{\text{reg}}$, respectively. Through this joint entropy-optimized training framework, we obtain a compact yet high-fidelity hierarchical 4D Gaussian representation, enabling efficient volumetric video compression for storage and transmission.

\begin{figure*}[t]
\centering
\includegraphics[width=\linewidth]{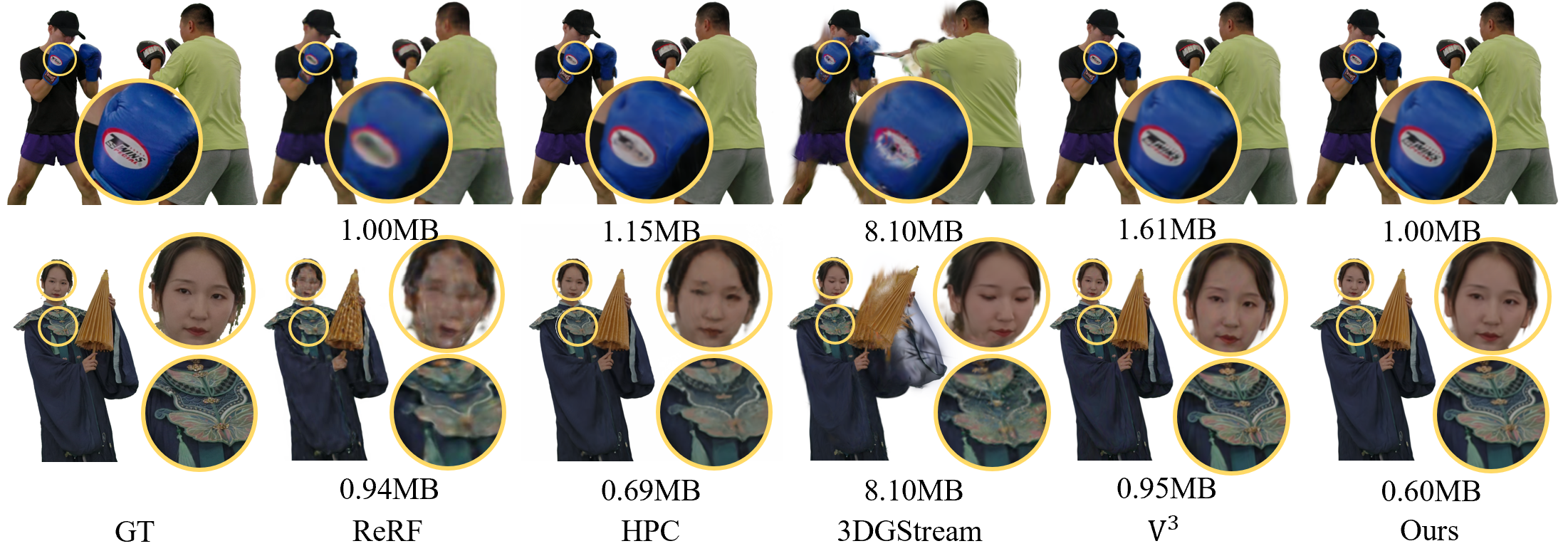}
\vspace{-2em}
\caption{Qualitative comparison on our 4DGCPro  and HiFi4G \cite{hifi4g} datasets against ReRF \cite{rerf}, HPC \cite{zheng2024hpc}, 3DGStream \cite{sun20243dgstream} and $\text{V}^{3}$ \cite{wang2024v}.}
\vspace{-2em}
\label{fig:Comparison}
\end{figure*}

\textbf{Efficient Progressive Bitstream Generation.} Once the training is completed, we explicitly separate Gaussians at different levels and implement differential quantization for Gaussian attributes, where we employ uint16 or uint32 precision for position information due to its higher sensitivity to errors while using uint8 for all other attributes. Since each Gaussian parameter contains multiple channels, we carefully flatten each feature channel into a separate 2D single-channel image while strictly preserving the 2D spatial continuity. The flattened feature images are systematically arranged into temporal sequences by aligning same-group, same-level, and same-channel features.
These sequences are then compressed using an H.264 video encoder, enabling scalable real-time decoding and rendering via hardware video codecs and shaders.
These bitstreams are transmitted collectively, enabling clients to selectively receive and decode different quality levels for adaptive rendering, thereby supporting smooth quality transitions, flexible viewing experiences, and real-time presentation.

\section{Experiments}
To comprehensively evaluate our method, we conducted experiments on two kind of distinct datasets: (1) the N3DV dataset \cite{li2022neural} featuring subtle motions with background, and (2) the HiFi4G dataset \cite{hifi4g} containing complex motions without background. We additionally captured a new dataset using 81 synchronized Z-CAM cinema cameras 3840$\times$2160, recording diverse performances including  dance, sports, and instrument playing. Our dataset contains not only solo performances but also multi-person interactions, which poses higher demands on the modeling method. We further introduce the detailed experimental settings in Sec. \ref{sec:appendixa}. 

\subsection{Comparison}
\begin{table}[t]
\centering
\setlength{\tabcolsep}{7pt}
\caption{Quantitative comparison on our 4DGCPro, HiFi4G \cite{hifi4g} and N3DV \cite{li2022neural} datasets. Our method achieves the best rendering quality against other methods, achieving a progressive rendering results within one single model.}
\vspace{-0.5em}
\label{tab:main_result}
\scalebox{0.85}{
\begin{tabular}{c|ccc|ccc|ccc}
\hline
\multirow{3}{*}{Method} & \multicolumn{3}{c|}{4DGCPro}                                                                                      & \multicolumn{3}{c|}{HiFi4G\cite{hifi4g}}                                                                                       & \multicolumn{3}{c}{N3DV\cite{li2022neural}}                                                                                         \\ \cline{2-10} 
                        & \begin{tabular}[c]{@{}c@{}}PSNR\\ (dB)\end{tabular}$\uparrow$ & SSIM$\uparrow$  & \begin{tabular}[c]{@{}c@{}}Size\\ (MB)\end{tabular}$\downarrow$ & \begin{tabular}[c]{@{}c@{}}PSNR\\ (dB)\end{tabular}$\uparrow$ & SSIM$\uparrow$  & \begin{tabular}[c]{@{}c@{}}Size\\ (MB)\end{tabular}$\downarrow$ & \begin{tabular}[c]{@{}c@{}}PSNR\\ (dB)\end{tabular}$\uparrow$ & SSIM$\uparrow$  & \begin{tabular}[c]{@{}c@{}}Size\\ (MB)\end{tabular}$\downarrow$ \\ \hline
ReRF\cite{rerf}                    & 27.57                                               & 0.947 & 1.70                                                & 30.30                                               & 0.977 & 0.97                                                & 29.71                                               & 0.918 & 0.77                                                \\
HPC\cite{zheng2024hpc}                     & 27.68                                               & 0.948 & 1.08                                                & 34.14                                               & 0.987 & 0.72                                                & -                                                   & -     & -                                                   \\
3DGStream\cite{sun20243dgstream}               & 21.08                                               & 0.837 & 8.1                                                 & 21.02                                               & 0.946 & 8.1                                                 & 31.54                                               & 0.942 & 8.10                                                 \\
4DGC\cite{hu20254dgcrateaware4dgaussian}                    & 21.48                                               & 0.850 & 0.97                                                & 21.05                                               & 0.946 & 0.94                                                & \underline{31.58}                                               & \underline{0.943} & 0.50                                                \\
HiCoM\cite{hicom2024}                   & 24.65                                               & 0.926 & 2.61                                                & 29.37                                               & 0.968 & 1.94                                                & 31.17                                               & 0.939 & 0.70                                                 \\
$\text{V}^{3}$\cite{wang2024v}                 & 28.11                                               & 0.955 & 1.60                                                & \underline{36.26}                                               & \underline{0.994} & 0.92                                                & -                                                   & -     & -                                                   \\ \hline
Ours(High)              & \textbf{29.47}                                               & \textbf{0.963} & 1.31                                                & \textbf{36.38}                                               & \textbf{0.995} & 0.75                                                & \textbf{31.64}                                               & \textbf{0.944} & 0.64                                                \\
Ours(Mid)            & \underline{28.68}                                               & \underline{0.958} & \underline{0.66}                                                & 35.48                                               & 0.991 & \underline{0.37}                                                & 31.14                                               & 0.938 & \underline{0.43}                                                \\
Ours(Low)               & 27.69                                               & 0.952 & \textbf{0.33}                                               & 34.62                                               & 0.988 & \textbf{0.19}                                                & 30.68                                               & 0.926 & \textbf{0.21}                                                \\ \hline
\end{tabular}
}
\vspace{-1em}
\end{table}
\begin{figure*}[t]
\centering
\includegraphics[width=\linewidth]{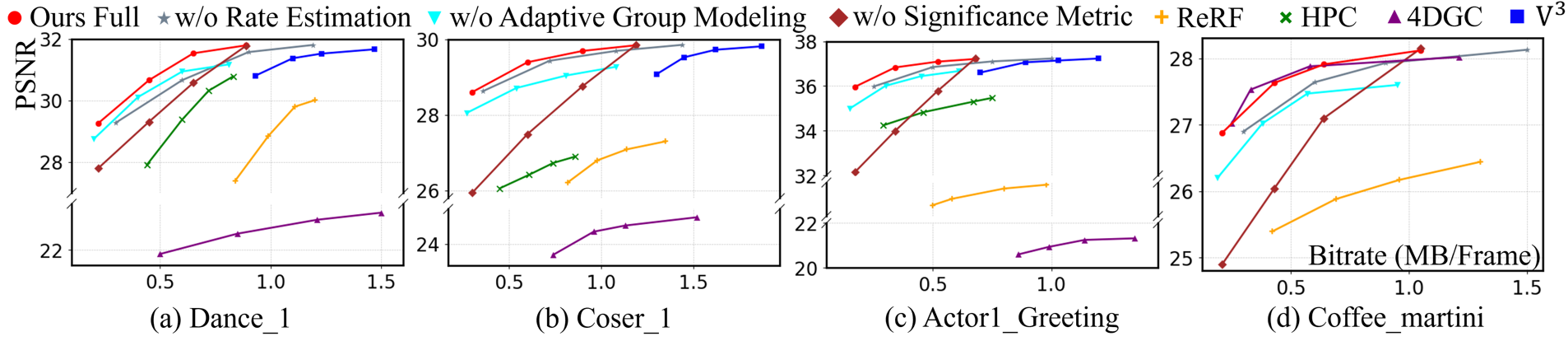}
\vspace{-1.5em}
\caption{Rate-distortion curves across various datasets. Rate-distortion curves not only illustrate the superiority of our method compared to the multiple-bitrate approaches ReRF \cite{rerf}, HPC \cite{zheng2024hpc}, 4DGC \cite{hu20254dgcrateaware4dgaussian}, and $\text{V}^{3}$ \cite{wang2024v}, but also demonstrate the efficiency of various components within our method.}
\vspace{-1em}
\label{fig:RD}
\end{figure*}
\begin{table}[t]
\centering
\setlength{\tabcolsep}{5.8pt}
\caption{The BD-PSNR results of our 4DGCPro, HPC \cite{zheng2024hpc}, 4DGC \cite{hu20254dgcrateaware4dgaussian} and $\text{V}^{3}$ \cite{wang2024v} when compared with ReRF \cite{rerf} on different datasets.}
\vspace{-0.5em}
\label{tab:compare_bdpsnr}
\scalebox{0.85}{
\begin{tabular}{c|cccc|cccc|cccc}
\hline
Dataset     & \multicolumn{4}{c|}{4DGCPro}  & \multicolumn{4}{c|}{HiFi4G\cite{hifi4g}}   & \multicolumn{4}{c}{N3DV\cite{li2022neural}}    \\ \hline
Method      & HPC  & 4DGC  & $\text{V}^{3}$ & Ours & HPC  & 4DGC  & $\text{V}^{3}$ & Ours & HPC & 4DGC & $\text{V}^{3}$ & Ours \\ \hline
BD-PSNR(dB)$\uparrow$ & 3.42 & -6.15 & 1.90    & \textbf{4.20} & 5.84 & -9.10 & 7.19    & \textbf{7.87} & -   & 1.99 & -       & \textbf{2.07} \\ \hline
\end{tabular}
}
\vspace{-0.5em}
\end{table}

\begin{table}[h]
\centering
\setlength{\tabcolsep}{9.1pt}
\caption{Complexity comparison of our method with dynamic scene compression methods, ReRF \cite{rerf}, HPC \cite{zheng2024hpc}, 4DGC \cite{hu20254dgcrateaware4dgaussian} and $\text{V}^{3}$ \cite{wang2024v} on 4DGCPro dataset.}
\vspace{-0.5em}
\label{tab:time}
\scalebox{0.85}{
\begin{tabular}{c|cccccc|ccc}
\hline
\multirow{2}{*}{Time} & \multirow{2}{*}{ReRF\cite{rerf}} & \multirow{2}{*}{4DGC\cite{hu20254dgcrateaware4dgaussian}} & \multirow{2}{*}{$\text{V}^{3}$\cite{wang2024v}} & \multicolumn{3}{c}{HPC\cite{zheng2024hpc}} & \multicolumn{3}{|c}{Ours} \\ \cline{5-10} 
                                                                     &                       &                       &                     & High  & Mid  & Low   & High   & Mid   & Low  \\ \hline 
Encode(ms)                                                             & 820                   & 2700                   & 390                 & 3300  & 2870    & 2420  & 408    & 205      & \textbf{102}  \\
Decode(ms)                                                             & 61                    & 94                    & 20                  & 121   & 103     & 90    & 29     & 19       & \textbf{12}   \\ 
Train(min)                                                             & 42.73                    & \textbf{0.83}                    & 0.97                  & 93   & 93     & 93    & 4.3     & 4.3      & 4.3   \\
Render(ms)                                                             & 52                    & 5.6                    & 2.8                  & 231   & 167     & 110    & 3.1     & 2.5       & \textbf{2.2}   \\\hline
\end{tabular}
}
\vspace{-1.5em}
\end{table}
To validate the effectiveness of the proposed method, we compare it against several SOTA approaches including NeRF-based methods ReRF \cite{rerf}, HPC \cite{zheng2024hpc} and 3DGS-based methods 3DGStream \cite{sun20243dgstream}, 4DGC \cite{hu20254dgcrateaware4dgaussian}, HiCoM \cite{hicom2024}, $\text{V}^{3}$ \cite{wang2024v}, presenting results in Fig. \ref{fig:Comparison}. It can be observed that due to the limitations of the finite neural representation of NeRF when dealing with complex motions, both ReRF \cite{rerf} and HPC \cite{zheng2024hpc} produce blurry results and over-smoothing. Meanwhile, 3DGStream \cite{sun20243dgstream} is limited to modeling only rigid motion of Gaussians and relies exclusively on the first frame as a universal reference across all frames. This leads to severe error accumulation over time, particularly in regions with motion, causing pronounced visual artifacts. Due to its inability to capture non-rigid deformations and significant displacements, the approach produces inconsistencies, including trajectory fragmentation and residual errors propagated from previous frames. Compared with $\text{V}^{3}$ \cite{wang2024v}, our method achieves better reconstruction quality while reducing the model bitrate, and it can render multi-quality reconstruction results from a single model. We further provide more demonstrations of our method in Sec. \ref{app:demonstrations}.

For quantitative comparison, as demonstrated in Tab. \ref{tab:main_result}, our method achieves superior performance compared to other approaches on diverse datasets. On the 4DGCPro dataset, our method achieves superior performance across all quality levels: the high-quality model attains the best PSNR (\textbf{29.47dB}) and SSIM (\textbf{0.963}) with compact size (\textbf{1.31MB}); the medium-quality version maintains high PSNR performance (\textbf{28.68dB}) with improved compression (\textbf{0.66MB}); while the low-quality configuration further reduces model size to \textbf{0.33MB} while retaining competitive quality (\textbf{27.69dB}). Notably, our framework supports multiple bitrates within a single model while outperforming baselines on all metrics. The rate-distortion superiority of our approach is further demonstrated in Fig.~\ref{fig:RD} and quantitatively validated through BD-PSNR measurements in Tab.~\ref{tab:compare_bdpsnr}. Our method achieves consistent RD improvements across all datasets and bitrates, with BD-PSNR gains of \textbf{4.20dB}, \textbf{7.87dB}, and \textbf{2.07dB} over ReRF on the 4DGCPro, HiFi4G, and N3DV datasets, respectively. These results significantly exceed those of other compared methods, including HPC (\textbf{3.42dB} on 4DGCPro) and $\text{V}^{3}$ (\textbf{1.90dB} on 4DGCPro). The superior RD performance demonstrates the effectiveness of our significance metric, adaptive grouping, and entropy modeling strategies.

As validated in Tab.~\ref{tab:time}, our method demonstrates exceptional computational efficiency across all quality levels. The medium-quality configuration achieves \textbf{19ms} decoding and \textbf{2.5ms} rendering per frame, enabling real-time performance at over \textbf{52 FPS}, while even the high-quality setting maintains practical efficiency with \textbf{29ms} decoding and \textbf{3.1ms} rendering. As shown in Tab.~\ref{tab:sup_t5}, our approach also delivers remarkable performance on lightweight devices: on mobile platforms, the complete pipeline requires only \textbf{43ms} for high-quality rendering, reduced to \textbf{39ms} and \textbf{34ms} for medium and low quality, demonstrating real-time decoding and rendering capability even under strict resource constraints.

These results collectively demonstrate that our method achieves the best trade-off between reconstruction quality, compression ratio, and computational efficiency among all compared approaches. The progressive coding capability further enhances practical applicability, enabling adaptive quality adjustment based on available computational resources and bandwidth constraints.

\subsection{Ablation Studies}
We conducted four ablation studies to evaluate the effectiveness of each component of our method. These experiments focus on the significance metric, motion-aware adaptive Gaussian grouping, the number of Gaussian layers, and joint entropy-optimized training. Using the full model as the baseline, we first ablated the components of the significance metric, including the weight parameter $\lambda_{\Psi}$. We then compared our adaptive grouping strategy against different fixed group lengths. The third experiment examines the impact of different number of Gaussian layers. Finally, we assessed various entropy modeling methods and underscored the importance of simulated quantization.

Tab. \ref{tab:ablation} presents ablation results for the significance metric and adaptive grouping strategy. The left section evaluates the significance metric for low-level Gaussians. Compared to our full model (\textbf{27.69dB}), using only opacity or volume leads to clear performance degradation (\textbf{26.71dB} and \textbf{25.83dB}, respectively). Simply multiplying these two factors also causes a significant PSNR drop of \textbf{-1.33dB}. Furthermore, improper weighting of \(\lambda_{\Psi}\) results in measurable PSNR reductions ranging from\textbf{ -0.18dB} to \textbf{-0.12dB}. The right section validates our motion-aware adaptive grouping approach. Even the best-performing fixed group size strategy shows consistent degradation, with a minimum BDBR of \textbf{8.11\%} and a maximum BD-PSNR reduction of \textbf{-0.25dB}, confirming the clear advantage of our adaptive grouping method in rate-distortion performance.

Table \ref{tab:ablation2} examines the impact of the number of Gaussian layers and the entropy modeling strategy. The left panel shows that deeper hierarchies (e.g., $L$=8) only slightly improve BD-PSNR (up to 0.09dB) at the cost of a noticeable increase in training time (\textbf{+1.2 min}). In contrast, shallower configurations (e.g., $L$=4) lead to a more substantial degradation in BD-PSNR (\textbf{-0.87dB}). The right panel highlights the importance of entropy modeling: omitting it ("w/o R-E") introduces redundancy, while removing hierarchical supervision ("w/o H-S") severely degrades the quality of low-level Gaussians, resulting in a BD-PSNR drop of \textbf{-2.89dB}. Moreover, the absence of simulated quantization (S-Q) leads to a \textbf{4.36\%} increase in BDBR, confirming its essential role in enhancing resilience to quantization errors. For Gaussian parameter modeling, using only KDE estimation ("Only KDE") achieves similar performance but prolongs training by \textbf{1.2 minutes} per frame, whereas assuming a universal Gaussian distribution ("Only Gaussian") causes training failures. Additional ablation studies are provided in Section \ref{sec:sup_ablation}.

\begin{table}[t]
    \centering
    \setlength{\tabcolsep}{4pt}
    \vspace{-0.5em}
    \caption{
    Ablation studies of our perceptually-weighted hierarchical Gaussian representation and adaptive Gaussian grouping.}
    \label{tab:ablation}
    \vspace{-0.5em}    
        \begin{minipage}{0.48\textwidth}
        \centering
        \scalebox{0.9}{
        \begin{tabular}{c|cc}
        \toprule
        Significance Metric & PSNR(dB)$\uparrow$ & Size(MB)$\downarrow$ \\ \midrule
        w/o Opacity                                                   & 26.71                                               & 0.39                                                \\
        w/o Volume                                                    & 25.83                                               & 0.38                                                \\
        Multiplication                   & 26.36                      & 0.33 \\
        $\lambda_{\Psi}$ = $2\times10^5$                                                           & 27.51                                               & 0.34                                                \\
        $\lambda_{\Psi}$ = $5\times10^4$                                                           & 27.57                                               & 0.35                                                \\
        Ours(Full)                                                    & \textbf{27.69}                                               & \textbf{0.33}                                                \\ \bottomrule
        \end{tabular}
        }
    \end{minipage}
    \hfill
    \begin{minipage}{0.48\textwidth}
        \centering
        \scalebox{0.9}{
        \begin{tabular}{c|cc}
        \toprule
        Group Size & BDBR(\%)$\downarrow$ & BD-PSNR(dB)$\uparrow$ \\ 
        \midrule
        1          & 48.37    & -0.96           \\
        5          & 11.81    & -0.25           \\
        10         & 16.34    & -0.32            \\
        15         & 14.95    & -0.33           \\
        20         & 11.42    & -0.34              \\
        25         & 8.11    & -0.31         \\
        \bottomrule
        \end{tabular}
        }
    \end{minipage}
    \hfill

\end{table}

\begin{table}[t]
    \centering
    \setlength{\tabcolsep}{4pt}
    \vspace{-0.5em}
    \caption{
    Ablation studies of the number of layers and end-to-end entropy-optimized training scheme.}
    \label{tab:ablation2}
    \vspace{-0.5em}  
    \begin{minipage}{0.48\textwidth}
        \centering
        \scalebox{0.9}{
        \begin{tabular}{c|cc}
        \toprule
         $L$    & BD-PSNR(dB)$\uparrow$ & Training Time(min)$\downarrow$ \\ 
        \midrule
        4        & -0.87    & 3.1        \\
        5                    & -0.38   & 3.5        \\
        6 & -    & 4.3        \\
        7               & 0.06    & 4.9         \\
        8              & 0.09    & 5.5        \\
        \bottomrule
        \end{tabular}
        }
    \end{minipage}
    \hfill
        \begin{minipage}{0.48\textwidth}
        \centering
        \scalebox{0.9}{
        \begin{tabular}{c|cc}
        \toprule
        Training     & BDBR(\%)$\downarrow$ & BD-PSNR(dB)$\uparrow$ \\ 
        \midrule
        w/o R-E        & 32.73    & -0.60        \\
        w/o H-S                    & 61.21   & -2.89        \\
        w/o S-Q & 4.36    & -0.1        \\
        Only KDE               & 0.58    & -0.02         \\
        Only Gaussian              & -    & -        \\
        \bottomrule
        \end{tabular}
        }
    \end{minipage}
    \hfill
\vspace{-1em}
\end{table}

\section{Discussion}
\textbf{Limitation.} Although 4DGCPro presents an innovative and efficient approach to progressive streaming of volumetric video, it has several limitations. First, the Gaussian optimization process suffers from prolonged training times (several minutes) due to hierarchical supervision which leads to repeated rendering passes. Accelerating this procedure remains an essential research objective. Second, our method relies on multi-view video input and faces challenges in sparse-view reconstruction, limiting its applicability in scenarios with insufficient camera coverage. Finally, the current framework underperforms in spatially extensive scenes, necessitating further exploration to enhance its scalability.

\textbf{Conclusion.} We present 4DGCPro, a novel hierarchical 4D Gaussian compression approach for progressive volumetric video streaming. Our framework accomplishes multiple bitrate control within a sigle model while supporting both real-time decoding and high-fidelity rendering on mobile platforms, even for sequences containing large motion displacements. Our approach begins by constructing a perceptually-weighted hierarchical Gaussian representation using the importance metric. We then model inter-frame Gaussians by rigid transformations and residual deformations, enhanced by a motion-aware adaptive Gaussian grouping strategy for efficient sequence-wide modeling. Furthermore, we introduce a joint entropy-optimized training and progressive coding framework, employing attribute-specific entropy modeling to ensure precise and efficient optimization. Thanks to its multiple bitrate capability, 4DGCPro enables progressive streaming and high-efficiency decoding/rendering across multiple quality levels, making it ideal for bandwidth-fluctuating scenarios. This work establishes a critical foundation for broader volumetric video adoption.

\section{Acknowledgements}
This work is supported by National Natural Science Foundation of China (62571322, 62431015, 62271308), STCSM (24ZR1432000, 24511106902, 24511106900, 22DZ2229005), 111 plan (BP0719010),  and State Key Laboratory of UHD Video and Audio Production and Presentation.

\bibliographystyle{Ref}  
\small
\bibliography{Reference}

\begin{thebibliography}{82}
\providecommand{\natexlab}[1]{#1}
\providecommand{\url}[1]{\texttt{#1}}
\expandafter\ifx\csname urlstyle\endcsname\relax
  \providecommand{\doi}[1]{doi: #1}\else
  \providecommand{\doi}{doi: \begingroup \urlstyle{rm}\Url}\fi

\bibitem[Akhtar et~al.(2022)Akhtar, Gao, Li, Li, Jia, and Liu]{9459517}
Akhtar , A., Gao , W., Li~, L., Li~, Z., Jia , W., \& Liu , S. (2022)
\newblock Video-based point cloud compression artifact removal.
\newblock \emph{IEEE Transactions on Multimedia} {\bfseries 24}:\penalty0 2866--2876.

\bibitem[Barron et~al.(2021)Barron, Mildenhall, Tancik, Hedman, Martin-Brualla, and Srinivasan]{barron2021mipnerf}
Barron , J.~T., Mildenhall , B., Tancik , M., Hedman , P., Martin-Brualla , R., \& Srinivasan , P.~P. (2021)
\newblock Mip-nerf: A multiscale representation for anti-aliasing neural radiance fields.
\newblock \emph{ICCV}

\bibitem[Barron et~al.(2022)Barron, Mildenhall, Verbin, Srinivasan, and Hedman]{barron2022mipnerf360}
Barron , J.~T., Mildenhall , B., Verbin , D., Srinivasan , P.~P., \& Hedman , P. (2022)
\newblock Mip-nerf 360: Unbounded anti-aliased neural radiance fields.
\newblock \emph{CVPR}

\bibitem[Barron et~al.(2023)Barron, Mildenhall, Verbin, Srinivasan, and Hedman]{barron2023zipnerf}
Barron , J.~T., Mildenhall , B., Verbin , D., Srinivasan , P.~P., \& Hedman , P. (2023)
\newblock Zip-nerf: Anti-aliased grid-based neural radiance fields.
\newblock \emph{ICCV}

\bibitem[Cao and Johnson(2023)]{hexplane_2023_CVPR}
Cao , A. \& Johnson , J. (2023)
\newblock Hexplane: A fast representation for dynamic scenes. In
\newblock \emph{CVPR}
\newblock pages 130--141.

\bibitem[Charatan et~al.(2023)Charatan, Li, Tagliasacchi, and Sitzmann]{charatan23pixelsplat}
Charatan , D., Li~, S., Tagliasacchi , A., \& Sitzmann , V. (2023)
\newblock pixelsplat: 3d gaussian splats from image pairs for scalable generalizable 3d reconstruction. In
\newblock \emph{arXiv}

\bibitem[Chen and Lee(2023)]{DBARF_Chen_2023_CVPR}
Chen , Y. \& Lee , G.~H. (2023)
\newblock Dbarf: Deep bundle-adjusting generalizable neural radiance fields. In
\newblock \emph{CVPR}
\newblock pages 24--34.

\bibitem[Dai et~al.(2017)Dai, Nie{\ss}ner, Zollh{\"o}fer, Izadi, and Theobalt]{dai2017bundlefusion}
Dai , A., Nie{\ss}ner , M., Zollh{\"o}fer , M., Izadi , S., \& Theobalt , C. (2017)
\newblock Bundlefusion: Real-time globally consistent 3d reconstruction using on-the-fly surface reintegration.
\newblock \emph{ACM Transactions on Graphics (ToG)} {\bfseries 36}\penalty0 (4):\penalty0 1.

\bibitem[Deng and Tartaglione(2023)]{cut}
Deng , C.~L. \& Tartaglione , E. (2023)
\newblock Compressing explicit voxel grid representations: fast nerfs become also small. In
\newblock \emph{Proceedings of the IEEE/CVF Winter Conference on Applications of Computer Vision}
\newblock pages 1236--1245.

\bibitem[Du et~al.(2021)Du, Zhang, Yu, Tenenbaum, and Wu]{NeuralRadianceFlow}
Du~, Y., Zhang , Y., Yu~, H.-X., Tenenbaum , J.~B., \& Wu~, J. (2021)
\newblock Neural radiance flow for 4d view synthesis and video processing. In
\newblock \emph{Proceedings of the IEEE/CVF International Conference on Computer Vision}

\bibitem[Fan et~al.(2024)Fan, Wang, Wen, Zhu, Xu, and Wang]{fan2023lightgaussian}
Fan , Z., Wang , K., Wen , K., Zhu , Z., Xu~, D., \& Wang , Z. (2024)
\newblock Lightgaussian: Unbounded 3d gaussian compression with 15x reduction and 200+ fps. In
\newblock \emph{Proc. Advances in Neural Information Processing Systems (NeurIPS)}

\bibitem[Fang et~al.(2022)Fang, Yi, Wang, Xie, Zhang, Liu, Nießner, and Tian]{tineuvox}
Fang , J., Yi~, T., Wang , X., Xie , L., Zhang , X., Liu , W., Nießner , M., \& Tian , Q. (2022)
\newblock Fast dynamic radiance fields with time-aware neural voxels. In
\newblock \emph{SIGGRAPH Asia 2022 Conference Papers}
\newblock ACM.

\bibitem[Feng et~al.(2024)Feng, Chen, Fu, Liao, Wang, Liu, Pei, Li, Zhang, and Dai]{feng2024flashgsefficient3dgaussian}
Feng , G., Chen , S., Fu~, R., Liao , Z., Wang , Y., Liu , T., Pei , Z., Li~, H., Zhang , X., \& Dai , B.
\newblock 2024.

\bibitem[Fridovich-Keil et~al.(2023)Fridovich-Keil, Meanti, Warburg, Recht, and Kanazawa]{kplanes}
Fridovich-Keil , S., Meanti , G., Warburg , F.~R., Recht , B., \& Kanazawa , A. (2023)
\newblock K-planes: Explicit radiance fields in space, time, and appearance. In
\newblock \emph{CVPR}
\newblock pages 12479--12488.

\bibitem[Fu et~al.(2022)Fu, Li, Song, Gao, and Liu]{OctAttention}
Fu~, C., Li~, G., Song , R., Gao , W., \& Liu , S. (2022)
\newblock Octattention: Octree-based large-scale contexts model for point cloud compression. In
\newblock \emph{the AAAI Conference on Artificial Intelligence}
\newblock \emph{36}, pp. \penalty0 625--633.

\bibitem[Gao et~al.(2024{\natexlab{a}})Gao, Meng, Wen, Chen, and Zhang]{hicom2024}
Gao , Q., Meng , J., Wen , C., Chen , J., \& Zhang , J. (2024.
\newblock {\natexlab{a}}) Hicom: Hierarchical coherent motion for dynamic streamable scenes with 3d gaussian splatting. In
\newblock \emph{Advances in Neural Information Processing Systems (NeurIPS)}

\bibitem[Gao et~al.(2024{\natexlab{b}})Gao, Planche, Zheng, Choudhuri, Chen, and Wu]{gao20246dgsenhanceddirectionawaregaussian}
Gao , Z., Planche , B., Zheng , M., Choudhuri , A., Chen , T., \& Wu~, Z.
\newblock 2024, {\natexlab{b}}.

\bibitem[Girish et~al.(2024)Girish, Li, Mazumdar, Shrivastava, Luebke, and De~Mello]{NEURIPS2024_4c9477b9}
Girish , S., Li~, T., Mazumdar , A., Shrivastava , A., Luebke , D., \& De~Mello , S. (2024)
\newblock Queen: Quantized efficient encoding of dynamic gaussians for streaming free-viewpoint videos. In
\newblock A.~Globerson, L.~Mackey, D.~Belgrave, A.~Fan, U.~Paquet, J.~Tomczak, and C.~Zhang, (eds.),
\newblock \emph{Advances in Neural Information Processing Systems}
\newblock \emph{37}, pp. \penalty0 43435--43467. Curran Associates, Inc.

\bibitem[Guo et~al.(2024)Guo, gang Zhou, Li, Wang, and Li]{Guo2024Motionaware3G}
Guo , Z., Zhou , W., Li~, L., Wang , M., \& Li~, H. (2024)
\newblock Motion-aware 3d gaussian splatting for efficient dynamic scene reconstruction.
\newblock \emph{ArXiv} {\bfseries abs/2403.11447}.

\bibitem[Hu et~al.(2025{\natexlab{a}})Hu, He, Zhong, Lu, Zhang, Zhai, and Wang]{10960681}
Hu~, Q., He~, Q., Zhong , H., Lu~, G., Zhang , X., Zhai , G., \& Wang , Y. (2025.
\newblock {\natexlab{a}})
\newblock Varfvv: View-adaptive real-time interactive free-view video streaming with edge computing.
\newblock \emph{IEEE Journal on Selected Areas in Communications} pages 1--1.

\bibitem[Hu et~al.(2025{\natexlab{b}})Hu, Zheng, Zhong, Fu, Song, XiaoyunZhang, Zhai, and Wang]{hu20254dgcrateaware4dgaussian}
Hu~, Q., Zheng , Z., Zhong , H., Fu~, S., Song , L., XiaoyunZhang , Zhai , G., \& Wang , Y. (2025.
\newblock {\natexlab{b}}) 4dgc: Rate-aware 4d gaussian compression for efficient streamable free-viewpoint video. In
\newblock \emph{CVPR}

\bibitem[Hu et~al.(2025{\natexlab{c}})Hu, Zhong, Zheng, Zhang, Cheng, Song, Zhai, and Wang]{vrvvc}
Hu~, Q., Zhong , H., Zheng , Z., Zhang , X., Cheng , Z., Song , L., Zhai , G., \& Wang , Y. (2025.
\newblock {\natexlab{c}})
\newblock Vrvvc: Variable-rate nerf-based volumetric video compression.
\newblock \emph{Proceedings of the AAAI Conference on Artificial Intelligence} {\bfseries 39}\penalty0 (4):\penalty0 3563--3571.

\bibitem[Huang et~al.(2024)Huang, Yu, Chen, Geiger, and Gao]{Huang2DGS2024}
Huang , B., Yu~, Z., Chen , A., Geiger , A., \& Gao , S. (2024)
\newblock 2d gaussian splatting for geometrically accurate radiance fields. In
\newblock \emph{SIGGRAPH 2024 Conference Papers}
\newblock Association for Computing Machinery.

\bibitem[Höllein et~al.(2024)Höllein, Božič, Zollhöfer, and Nießner]{höllein20243dgslmfastergaussiansplattingoptimization}
Höllein , L., Božič , A., Zollhöfer , M., \& Nießner , M.
\newblock 2024.

\bibitem[I\c{s}{\i}k et~al.(2023)I\c{s}{\i}k, Rünz, Georgopoulos, Khakhulin, Starck, Agapito, and Nießner]{HumanRF}
I\c{s}{\i}k , M., Rünz , M., Georgopoulos , M., Khakhulin , T., Starck , J., Agapito , L., \& Nießner , M. (2023)
\newblock Humanrf: High-fidelity neural radiance fields for humans in motion.
\newblock \emph{ACM Transactions on Graphics (TOG)} {\bfseries 42}\penalty0 (4).

\bibitem[Jiang et~al.(2024{\natexlab{a}})Jiang, Shen, Hong, Guo, Wu, Zhang, Yu, and Xu]{jiang2024robust}
Jiang , Y., Shen , Z., Hong , Y., Guo , C., Wu~, Y., Zhang , Y., Yu~, J., \& Xu~, L. (2024.
\newblock {\natexlab{a}})
\newblock Robust dual gaussian splatting for immersive human-centric volumetric videos.
\newblock \emph{arXiv preprint arXiv:2409.08353}

\bibitem[Jiang et~al.(2024{\natexlab{b}})Jiang, Shen, Wang, Su, Hong, Zhang, Yu, and Xu]{hifi4g}
Jiang , Y., Shen , Z., Wang , P., Su~, Z., Hong , Y., Zhang , Y., Yu~, J., \& Xu~, L. (2024.
\newblock {\natexlab{b}}) Hifi4g: High-fidelity human performance rendering via compact gaussian splatting. In
\newblock \emph{CVPR}
\newblock pages 19734--19745.

\bibitem[Kerbl et~al.(2023)Kerbl, Kopanas, Leimk{\"u}hler, and Drettakis]{kerbl3Dgaussians}
Kerbl , B., Kopanas , G., Leimk{\"u}hler , T., \& Drettakis , G. (2023)
\newblock 3d gaussian splatting for real-time radiance field rendering.
\newblock \emph{ACM Transactions on Graphics} {\bfseries 42}\penalty0 (4).

\bibitem[Levoy and Hanrahan(2023)]{LightFieldRendering}
Levoy , M. \& Hanrahan , P.
\newblock \emph{Light Field Rendering}.
\newblock Association for Computing Machinery, New York, NY, USA, 2023.

\bibitem[Li et~al.(2020)Li, Li, Zakharchenko, Chen, and Li]{8786879}
Li~, L., Li~, Z., Zakharchenko , V., Chen , J., \& Li~, H. (2020)
\newblock Advanced 3d motion prediction for video-based dynamic point cloud compression.
\newblock \emph{IEEE Transactions on Image Processing} {\bfseries 29}:\penalty0 289--302.

\bibitem[Li et~al.(2022{\natexlab{a}})Li, Shen, Wang, Shen, and Tan]{streaming}
Li~, L., Shen , Z., Wang , Z., Shen , L., \& Tan , P. (2022.
\newblock {\natexlab{a}})
\newblock Streaming radiance fields for 3d video synthesis.
\newblock \emph{Advances in Neural Information Processing Systems} {\bfseries 35}:\penalty0 13485--13498.

\bibitem[Li et~al.(2022{\natexlab{b}})Li, Slavcheva, Zollhoefer, Green, Lassner, Kim, Schmidt, Lovegrove, Goesele, Newcombe, et~al.]{li2022neural}
Li~, T., Slavcheva , M., Zollhoefer , M., Green , S., Lassner , C., Kim , C., Schmidt , T., Lovegrove , S., Goesele , M., Newcombe , R., \& others  (2022.
\newblock {\natexlab{b}}) Neural 3d video synthesis from multi-view video. In
\newblock \emph{CVPR}
\newblock pages 5521--5531.

\bibitem[Li et~al.(2021)Li, Niklaus, Snavely, and Wang]{9578364}
Li~, Z., Niklaus , S., Snavely , N., \& Wang , O. (2021)
\newblock Neural scene flow fields for space-time view synthesis of dynamic scenes. In
\newblock \emph{CVPR}
\newblock pages 6494--6504.

\bibitem[Li et~al.(2023)Li, Wang, Cole, Tucker, and Snavely]{Li_2023_CVPR}
Li~, Z., Wang , Q., Cole , F., Tucker , R., \& Snavely , N. (2023)
\newblock Dynibar: Neural dynamic image-based rendering. In
\newblock \emph{CVPR}
\newblock pages 4273--4284.

\bibitem[Li et~al.(2024)Li, Chen, Li, and Xu]{Li_STG_2024_CVPR}
Li~, Z., Chen , Z., Li~, Z., \& Xu~, Y. (2024)
\newblock Spacetime gaussian feature splatting for real-time dynamic view synthesis. In
\newblock \emph{CVPR}
\newblock pages 8508--8520.

\bibitem[Liang and Liang(2022)]{10.1145/3512527.3531423}
Liang , Z. \& Liang , F. (2022)
\newblock Transpcc: Towards deep point cloud compression via transformers. In
\newblock \emph{Proceedings of the 2022 International Conference on Multimedia Retrieval}
\newblock page 1–5, New York, NY, USA: Association for Computing Machinery.

\bibitem[Loza et~al.(2006)Loza, Mihaylova, Canagarajah, and Bull]{dssim}
Loza , A., Mihaylova , L., Canagarajah , N., \& Bull , D. (2006)
\newblock Structural similarity-based object tracking in video sequences. In
\newblock \emph{2006 9th International Conference on Information Fusion}
\newblock pages 1--6.

\bibitem[Lu et~al.(2024)Lu, Yu, Xu, Xiangli, Wang, Lin, and Dai]{scaffoldgs}
Lu~, T., Yu~, M., Xu~, L., Xiangli , Y., Wang , L., Lin , D., \& Dai , B. (2024)
\newblock Scaffold-gs: Structured 3d gaussians for view-adaptive rendering. In
\newblock \emph{CVPR}
\newblock pages 20654--20664.

\bibitem[Luiten et~al.(2024)Luiten, Kopanas, Leibe, and Ramanan]{luiten2023dynamic}
Luiten , J., Kopanas , G., Leibe , B., \& Ramanan , D. (2024)
\newblock Dynamic 3d gaussians: Tracking by persistent dynamic view synthesis. In
\newblock \emph{3DV}

\bibitem[Martin-Brualla et~al.(2021)Martin-Brualla, Radwan, Sajjadi, Barron, Dosovitskiy, and Duckworth]{martinbrualla2020nerfw}
Martin-Brualla , R., Radwan , N., Sajjadi , M. S.~M., Barron , J.~T., Dosovitskiy , A., \& Duckworth , D. (2021)
\newblock {NeRF in the Wild: Neural Radiance Fields for Unconstrained Photo Collections}. In
\newblock \emph{CVPR}

\bibitem[Mildenhall et~al.(2021)Mildenhall, Srinivasan, Tancik, Barron, Ramamoorthi, and Ng]{nerf}
Mildenhall , B., Srinivasan , P.~P., Tancik , M., Barron , J.~T., Ramamoorthi , R., \& Ng~, R. (2021)
\newblock Nerf: Representing scenes as neural radiance fields for view synthesis.
\newblock \emph{Communications of the ACM} {\bfseries 65}\penalty0 (1):\penalty0 99--106.

\bibitem[M\"uller et~al.(2022)M\"uller, Evans, Schied, and Keller]{instant-ngp}
M\"uller , T., Evans , A., Schied , C., \& Keller , A. (2022)
\newblock Instant neural graphics primitives with a multiresolution hash encoding.
\newblock \emph{ACM Trans. Graph.} {\bfseries 41}\penalty0 (4):\penalty0 102:1--102:15.

\bibitem[Nadenau et~al.(2003)Nadenau, Reichel, and Kunt]{Wavelet-based}
Nadenau , M., Reichel , J., \& Kunt , M. (2003)
\newblock Wavelet-based color image compression: Exploiting the contrast sensitivity function.
\newblock \emph{IEEE transactions on image processing : a publication of the IEEE Signal Processing Society} {\bfseries 12}:\penalty0 58--70.

\bibitem[Navaneet et~al.(2024)Navaneet, Meibodi, Koohpayegani, and Pirsiavash]{navaneet2023compact3d}
Navaneet , K., Meibodi , K.~P., Koohpayegani , S.~A., \& Pirsiavash , H. (2024)
\newblock Compgs: Smaller and faster gaussian splatting with vector quantization.
\newblock \emph{ECCV}

\bibitem[Overbeck et~al.(2018)Overbeck, Erickson, Evangelakos, Pharr, and Debevec]{panoramiclightfieldsystem}
Overbeck , R.~S., Erickson , D., Evangelakos , D., Pharr , M., \& Debevec , P. (2018)
\newblock A system for acquiring, processing, and rendering panoramic light field stills for virtual reality.
\newblock \emph{ACM Trans. Graph.} {\bfseries 37}\penalty0 (6).

\bibitem[Park et~al.(2021)Park, Sinha, Barron, Bouaziz, Goldman, Seitz, and Martin-Brualla]{Nerfies}
Park , K., Sinha , U., Barron , J.~T., Bouaziz , S., Goldman , D.~B., Seitz , S.~M., \& Martin-Brualla , R. (2021)
\newblock Nerfies: Deformable neural radiance fields. In
\newblock \emph{ICCV (ICCV)}
\newblock pages 5865--5874.

\bibitem[Park et~al.(2023{\natexlab{a}})Park, Henzler, Mildenhall, and Barron]{park2023camp}
Park , K., Henzler , P., Mildenhall , B., \& Barron , R. (2023.
\newblock {\natexlab{a}})
\newblock Camp: Camera preconditioning for neural radiance fields.
\newblock \emph{ACM Trans. Graph.}

\bibitem[Park et~al.(2023{\natexlab{b}})Park, Son, Jang, Ahn, Kim, and Kang]{Park2023TemporalII}
Park , S., Son , M., Jang , S., Ahn , Y.~C., Kim , J.-Y., \& Kang , N. (2023.
\newblock {\natexlab{b}})
\newblock Temporal interpolation is all you need for dynamic neural radiance fields.
\newblock \emph{CVPR} pages 4212--4221.

\bibitem[Peng et~al.(2023)Peng, Yan, Shuai, Bao, and Zhou]{peng2023representing}
Peng , S., Yan , Y., Shuai , Q., Bao , H., \& Zhou , X. (2023)
\newblock Representing volumetric videos as dynamic mlp maps. In
\newblock \emph{CVPR}
\newblock pages 4252--4262.

\bibitem[Pumarola et~al.(2020)Pumarola, Corona, Pons-Moll, and Moreno-Noguer]{DNeRF}
Pumarola , A., Corona , E., Pons-Moll , G., \& Moreno-Noguer , F. (2020)
\newblock {D-NeRF: Neural Radiance Fields for Dynamic Scenes}. In
\newblock \emph{CVPR}

\bibitem[Quach et~al.(2019)Quach, Valenzise, and Dufaux]{8803413}
Quach , M., Valenzise , G., \& Dufaux , F. (2019)
\newblock Learning convolutional transforms for lossy point cloud geometry compression. In
\newblock \emph{2019 IEEE International Conference on Image Processing (ICIP)}
\newblock pages 4320--4324.

\bibitem[Quach et~al.(2020)Quach, Valenzise, and Dufaux]{9287077}
Quach , M., Valenzise , G., \& Dufaux , F. (2020)
\newblock Improved deep point cloud geometry compression. In
\newblock \emph{2020 IEEE 22nd International Workshop on Multimedia Signal Processing (MMSP)}
\newblock pages 1--6.

\bibitem[Rabich et~al.(2023)Rabich, Stotko, and Klein]{fpo++}
Rabich , S., Stotko , P., \& Klein , R. (2023)
\newblock Fpo++: Efficient encoding and rendering of dynamic neural radiance fields by analyzing and enhancing fourier plenoctrees.
\newblock \emph{arXiv preprint arXiv:2310.20710}

\bibitem[Reiser et~al.(2023)Reiser, Szeliski, Verbin, Srinivasan, Mildenhall, Geiger, Barron, and Hedman]{merf}
Reiser , C., Szeliski , R., Verbin , D., Srinivasan , P., Mildenhall , B., Geiger , A., Barron , J., \& Hedman , P. (2023)
\newblock Merf: Memory-efficient radiance fields for real-time view synthesis in unbounded scenes.
\newblock \emph{ACM Transactions on Graphics (TOG)} {\bfseries 42}\penalty0 (4):\penalty0 1--12.

\bibitem[Rho et~al.(2023)Rho, Lee, Nam, Lee, Ko, and Park]{miniwave}
Rho , D., Lee , B., Nam , S., Lee , J.~C., Ko~, J.~H., \& Park , E. (2023)
\newblock Masked wavelet representation for compact neural radiance fields. In
\newblock \emph{CVPR}
\newblock pages 20680--20690.

\bibitem[Schnabel and Klein(2006)]{10.5555/2386388.2386404}
Schnabel , R. \& Klein , R. (2006)
\newblock Octree-based point-cloud compression. In
\newblock \emph{Proceedings of the 3rd Eurographics / IEEE VGTC Conference on Point-Based Graphics}
\newblock page 111–121, Goslar, DEU: Eurographics Association.

\bibitem[Schwarz et~al.(2019)Schwarz, Preda, Baroncini, Budagavi, Cesar, Chou, Cohen, Krivokuća, Lasserre, Li, Llach, Mammou, Mekuria, Nakagami, Siahaan, Tabatabai, Tourapis, and Zakharchenko]{8571288}
Schwarz , S., Preda , M., Baroncini , V., Budagavi , M., Cesar , P., Chou , P.~A., Cohen , R.~A., Krivokuća , M., Lasserre , S., Li~, Z., Llach , J., Mammou , K., Mekuria , R., Nakagami , O., Siahaan , E., Tabatabai , A., Tourapis , A.~M., \& Zakharchenko , V. (2019)
\newblock Emerging mpeg standards for point cloud compression.
\newblock \emph{IEEE Journal on Emerging and Selected Topics in Circuits and Systems} {\bfseries 9}\penalty0 (1):\penalty0 133--148.

\bibitem[Schönberger and Frahm(2016)]{schoenberger2016sfm}
Schönberger , J.~L. \& Frahm , J.-M. (2016)
\newblock Structure-from-motion revisited. In
\newblock \emph{2016 IEEE Conference on Computer Vision and Pattern Recognition (CVPR)}
\newblock pages 4104--4113.

\bibitem[Shao et~al.(2023)Shao, Zheng, Tu, Liu, Zhang, and Liu]{tensor4d}
Shao , R., Zheng , Z., Tu~, H., Liu , B., Zhang , H., \& Liu , Y. (2023)
\newblock Tensor4d: Efficient neural 4d decomposition for high-fidelity dynamic reconstruction and rendering. In
\newblock \emph{CVPR}
\newblock pages 16632--16642.

\bibitem[Song et~al.(2023)Song, Chen, Li, Chen, Chen, Yuan, Xu, and Geiger]{nerfplayer}
Song , L., Chen , A., Li~, Z., Chen , Z., Chen , L., Yuan , J., Xu~, Y., \& Geiger , A. (2023)
\newblock Nerfplayer: A streamable dynamic scene representation with decomposed neural radiance fields.
\newblock \emph{IEEE Transactions on Visualization and Computer Graphics} {\bfseries 29}\penalty0 (5):\penalty0 2732--2742.

\bibitem[Su et~al.(2020)Su, Xu, Zheng, Yu, Liu, and Fang]{RobustFusion}
Su~, Z., Xu~, L., Zheng , Z., Yu~, T., Liu , Y., \& Fang , L. (2020)
\newblock Robustfusion: Human volumetric capture with data-driven visual cues using a rgbd camera. In
\newblock \emph{Computer Vision – ECCV 2020: 16th European Conference, Glasgow, UK, August 23–28, 2020, Proceedings, Part IV}
\newblock page 246–264, Berlin, Heidelberg: Springer-Verlag.

\bibitem[Sun et~al.(2024)Sun, Jiao, Li, Zhang, Zhao, and Xing]{sun20243dgstream}
Sun , J., Jiao , H., Li~, G., Zhang , Z., Zhao , L., \& Xing , W. (2024)
\newblock 3dgstream: On-the-fly training of 3d gaussians for efficient streaming of photo-realistic free-viewpoint videos. In
\newblock \emph{CVPR}
\newblock pages 20675--20685.

\bibitem[Thanou et~al.(2016)Thanou, Chou, and Frossard]{7405340}
Thanou , D., Chou , P.~A., \& Frossard , P. (2016)
\newblock Graph-based compression of dynamic 3d point cloud sequences.
\newblock \emph{IEEE Transactions on Image Processing} {\bfseries 25}\penalty0 (4):\penalty0 1765--1778.

\bibitem[Wang et~al.(2023{\natexlab{a}})Wang, Tan, Li, Tian, Song, and Liu]{10377294}
Wang , F., Tan , S., Li~, X., Tian , Z., Song , Y., \& Liu , H. (2023.
\newblock {\natexlab{a}}) Mixed neural voxels for fast multi-view video synthesis. In
\newblock \emph{ICCV}
\newblock pages 19649--19659.

\bibitem[Wang et~al.(2021)Wang, Zhu, Liu, and Ma]{9321375}
Wang , J., Zhu , H., Liu , H., \& Ma~, Z. (2021)
\newblock Lossy point cloud geometry compression via end-to-end learning.
\newblock \emph{IEEE Transactions on Circuits and Systems for Video Technology} {\bfseries 31}\penalty0 (12):\penalty0 4909--4923.

\bibitem[Wang et~al.(2022)Wang, Zhang, Liu, Zhao, Zhang, Zhang, Wu, Yu, and Xu]{9879176}
Wang , L., Zhang , J., Liu , X., Zhao , F., Zhang , Y., Zhang , Y., Wu~, M., Yu~, J., \& Xu~, L. (2022)
\newblock Fourier plenoctrees for dynamic radiance field rendering in real-time. In
\newblock \emph{CVPR}
\newblock pages 13514--13524.

\bibitem[Wang et~al.(2023{\natexlab{b}})Wang, Hu, He, Wang, Yu, Tuytelaars, Xu, and Wu]{rerf}
Wang , L., Hu~, Q., He~, Q., Wang , Z., Yu~, J., Tuytelaars , T., Xu~, L., \& Wu~, M. (2023.
\newblock {\natexlab{b}}) Neural residual radiance fields for streamably free-viewpoint videos. In
\newblock \emph{CVPR}
\newblock pages 76--87.

\bibitem[Wang et~al.(2023{\natexlab{c}})Wang, Yao, Guo, Zhang, Hu, Yu, Xu, and Wu]{videorf}
Wang , L., Yao , K., Guo , C., Zhang , Z., Hu~, Q., Yu~, J., Xu~, L., \& Wu~, M.
\newblock 2023, {\natexlab{c}}.

\bibitem[Wang et~al.(2018)Wang, Zhang, Li, Fu, Liu, and Jiang]{Wang_2018_ECCV}
Wang , N., Zhang , Y., Li~, Z., Fu~, Y., Liu , W., \& Jiang , Y.-G. (2018)
\newblock Pixel2mesh: Generating 3d mesh models from single rgb images. In
\newblock \emph{Proceedings of the European Conference on Computer Vision (ECCV)}

\bibitem[Wang et~al.(2024{\natexlab{a}})Wang, Zhang, Wang, Yao, Xie, Yu, Wu, and Xu]{wang2024v}
Wang , P., Zhang , Z., Wang , L., Yao , K., Xie , S., Yu~, J., Wu~, M., \& Xu~, L. (2024.
\newblock {\natexlab{a}})
\newblock V\^{} 3: Viewing volumetric videos on mobiles via streamable 2d dynamic gaussians.
\newblock \emph{ACM Transactions on Graphics (TOG)} {\bfseries 43}\penalty0 (6):\penalty0 1--13.

\bibitem[Wang et~al.(2023{\natexlab{d}})Wang, Han, Habermann, Daniilidis, Theobalt, and Liu]{neus2}
Wang , Y., Han , Q., Habermann , M., Daniilidis , K., Theobalt , C., \& Liu , L. (2023.
\newblock {\natexlab{d}}) Neus2: Fast learning of neural implicit surfaces for multi-view reconstruction. In
\newblock \emph{ICCV}

\bibitem[Wang et~al.(2024{\natexlab{b}})Wang, Li, Guo, Yang, Kot, and Wen]{wang2024contextgs}
Wang , Y., Li~, Z., Guo , L., Yang , W., Kot , A.~C., \& Wen , B. (2024.
\newblock {\natexlab{b}})
\newblock Contextgs: Compact 3d gaussian splatting with anchor level context model.
\newblock \emph{arXiv preprint arXiv:2405.20721}

\bibitem[Wu et~al.(2024{\natexlab{a}})Wu, Yi, Fang, Xie, Zhang, Wei, Liu, Tian, and Wang]{Wu_2024_CVPR}
Wu~, G., Yi~, T., Fang , J., Xie , L., Zhang , X., Wei , W., Liu , W., Tian , Q., \& Wang , X. (2024.
\newblock {\natexlab{a}}) 4d gaussian splatting for real-time dynamic scene rendering. In
\newblock \emph{CVPR}
\newblock pages 20310--20320.

\bibitem[Wu et~al.(2024{\natexlab{b}})Wu, Wang, Kouros, and Tuytelaars]{tetrirf}
Wu~, M., Wang , Z., Kouros , G., \& Tuytelaars , T. (2024.
\newblock {\natexlab{b}}) Tetrirf: Temporal tri-plane radiance fields for efficient free-viewpoint video. In
\newblock \emph{CVPR}
\newblock pages 6487--6496.

\bibitem[Xu et~al.(2024)Xu, Peng, Lin, He, Sun, Shen, Bao, and Zhou]{xu20244k4d}
Xu~, Z., Peng , S., Lin , H., He~, G., Sun , J., Shen , Y., Bao , H., \& Zhou , X. (2024)
\newblock 4k4d: Real-time 4d view synthesis at 4k resolution. In
\newblock \emph{2024 IEEE/CVF Conference on Computer Vision and Pattern Recognition (CVPR)}
\newblock pages 20029--20040.

\bibitem[Yan et~al.(2024)Yan, Peng, Tang, and Wang]{yan20244d}
Yan , J., Peng , R., Tang , L., \& Wang , R. (2024)
\newblock 4d gaussian splatting with scale-aware residual field and adaptive optimization for real-time rendering of temporally complex dynamic scenes. In
\newblock \emph{ACM MM}
\newblock pages 7871--7880.

\bibitem[Yang et~al.(2023)Yang, Gao, Zhou, Jiao, Zhang, and Jin]{yang2023deformable3dgs}
Yang , Z., Gao , X., Zhou , W., Jiao , S., Zhang , Y., \& Jin , X. (2023)
\newblock Deformable 3d gaussians for high-fidelity monocular dynamic scene reconstruction.
\newblock \emph{arXiv preprint arXiv:2309.13101}

\bibitem[Yang et~al.(2024)Yang, Yang, Pan, and Zhang]{yang2023gs4d}
Yang , Z., Yang , H., Pan , Z., \& Zhang , L. (2024)
\newblock Real-time photorealistic dynamic scene representation and rendering with 4d gaussian splatting. In

\bibitem[Zhang et~al.(2021)Zhang, Liu, Ye, Zhao, Zhang, Wu, Zhang, Xu, and Yu]{zhang2021editable}
Zhang , J., Liu , X., Ye~, X., Zhao , F., Zhang , Y., Wu~, M., Zhang , Y., Xu~, L., \& Yu~, J. (2021)
\newblock Editable free-viewpoint video using a layered neural representation.
\newblock \emph{ACM Transactions on Graphics (TOG)} {\bfseries 40}\penalty0 (4):\penalty0 1--18.

\bibitem[Zhang(2012)]{kinectsensor}
Zhang , Z. (2012)
\newblock Microsoft kinect sensor and its effect.
\newblock \emph{IEEE MultiMedia} {\bfseries 19}\penalty0 (2):\penalty0 4–10.

\bibitem[Zheng et~al.(2024{\natexlab{a}})Zheng, Zhong, Hu, Zhang, Song, Zhang, and Wang]{zheng2024hpc}
Zheng , Z., Zhong , H., Hu~, Q., Zhang , X., Song , L., Zhang , Y., \& Wang , Y. (2024.
\newblock {\natexlab{a}}) Hpc: Hierarchical progressive coding framework for volumetric video. In
\newblock \emph{ACM MM}
\newblock page 7937–7946, New York, NY, USA: Association for Computing Machinery.

\bibitem[Zheng et~al.(2024{\natexlab{b}})Zheng, Zhong, Hu, Zhang, Song, Zhang, and Wang]{zheng2024jointrf}
Zheng , Z., Zhong , H., Hu~, Q., Zhang , X., Song , L., Zhang , Y., \& Wang , Y. (2024.
\newblock {\natexlab{b}}) Jointrf: End-to-end joint optimization for dynamic neural radiance field representation and compression. In
\newblock \emph{2024 IEEE International Conference on Image Processing (ICIP)}
\newblock pages 3292--3298.

\end{thebibliography}
\normalsize

\newpage

\appendix
This appendix provides additional material to supplement the main text. We will first introduce implementation details in Sec. \ref{sec:appendixa}. Then we provide additional experimental results in Sec. \ref{sec:appendixb}.
\begin{figure*}[t]
\centering
\includegraphics[width=\linewidth]{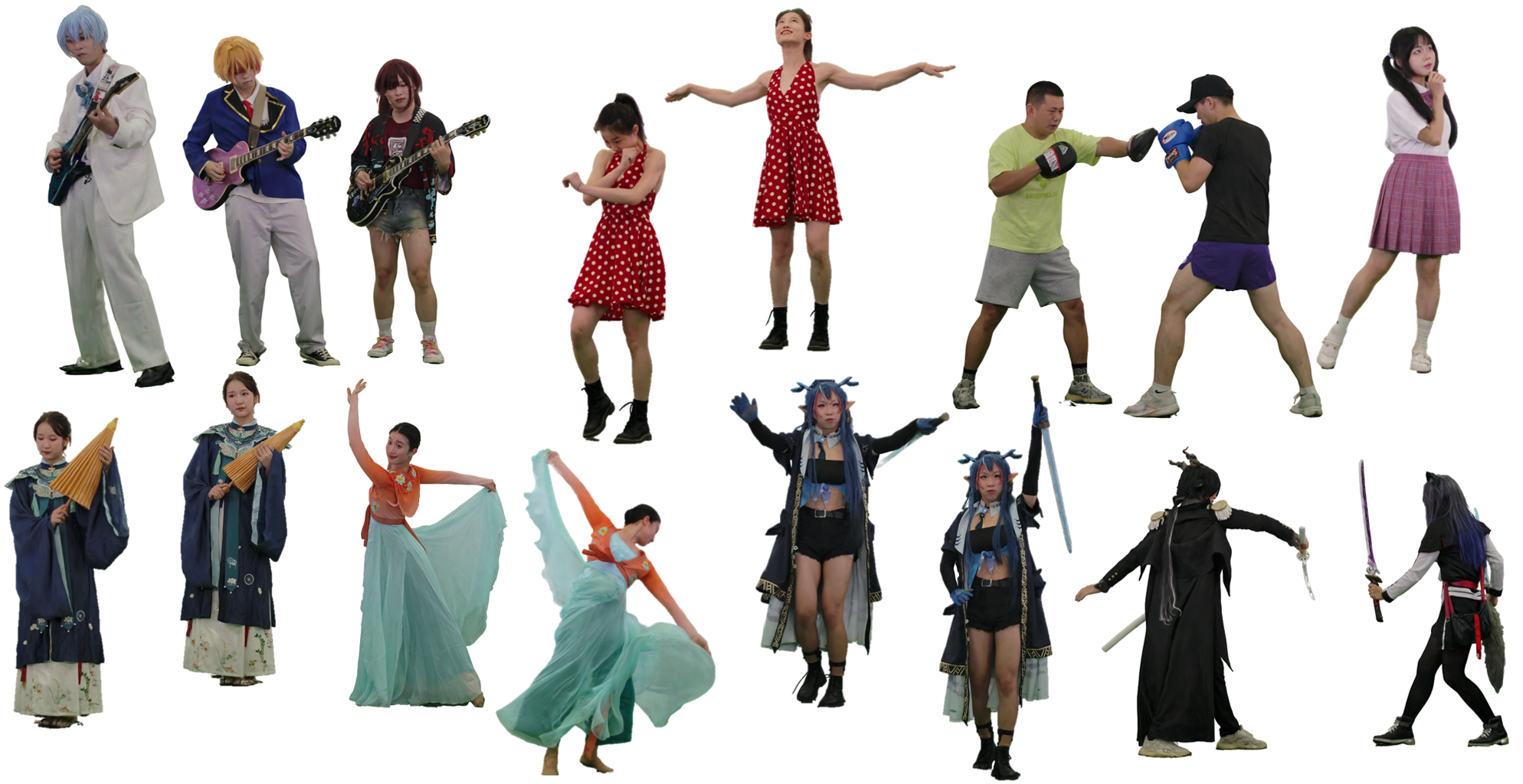}
\vspace{-1em}
\caption{Gallery of our results. 4DGCPro delivers real-time high-fidelity rendering of scenes across challenging motions, such as “playing instruments”, “dancing” and “playing sports”. }
\label{fig:gallery}
\end{figure*}
\begin{figure*}[t]
\centering
\includegraphics[width=\linewidth]{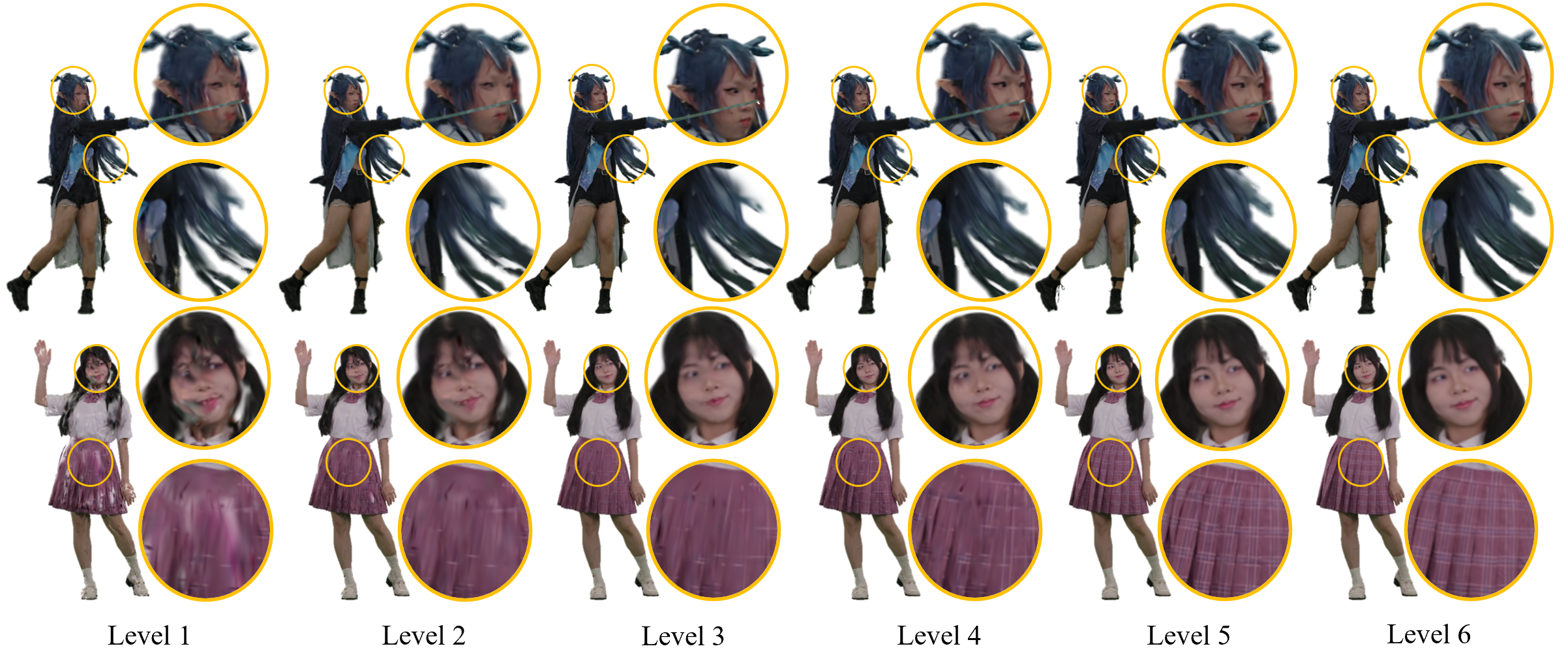}
\vspace{-1.5em}
\caption{Multi-bitrate results of our method under a single
bitstream.}
\vspace{-0.5em}
\label{fig:hierarchical}
\end{figure*}

\section{Implementation Details}
\label{sec:appendixa}
Our code is primarily based on the open-source codes of 3DGS \cite{kerbl3Dgaussians} and $\text{V}^{3}$ \cite{wang2024v} and is also inspired by 3DGStream \cite{sun20243dgstream} and 4DGC \cite{hu20254dgcrateaware4dgaussian}. In the initialization phase, due to the limitations of the NeuS2 \cite{neus2} method, it is challenging to obtain high-quality surface meshes for scenes with backgrounds, making it difficult to initialize Gaussians effectively. Therefore, on the N3DV dataset, we still initialize Gaussians based on the results of COLMAP. Additionally, we observed that in scenes with multiple interacting people, NeuS2 may occasionally fail to capture all individuals. To enhance the stability of our method, we train the NeuS2 network parameters of each keyframe to learn residuals from a pre-trained, known-correct NeuS2 network. During the pre-training phase of key frames, we first train for 12,000 steps under the setting of $\lambda_{\text{ssim}}=0.2$. In the Gaussian pruning phase, we remove 40\% of the Gaussians with lower opacity on HiFi4G dataset \cite{hifi4g} and 4DGCPro dataset but not remove any Gaussians on N3DV dataset \cite{li2022neural}. During the hierarchical process, we set $\lambda_{\Psi}=1\times10^5$ to ensure a balance between volume and opacity, and divide the Gaussians into $L = 6$ layers. Regarding the motion-aware adaptive Gaussian grouping, we have selected different $\tau_{\mu}$ values for different datasets: $\tau_{\mu} = 0.0025$ for 4DGCpro, $\tau_{\mu} = 0.001$ for HiFi4G, and $\tau_{\mu} = 0.01$ for N3DV. Then, we conduct end-to-end entropy-optimized training on keyframes for 1,500 steps with the supervision of $\lambda^l=\begin{cases}0.5/l, & l < L\\1, & l = L\end{cases}$, and $\lambda_{\text{rate\_key}}=1\times10^{-7}$. In the subsequent inter-frame Optimization., we set $\lambda_{\text{rate\_inter}}= 1\times10^{-4}$, $\lambda_{\text{reg}} = 1\times10^{-3}$, and train for 800 and 2,000 steps in the rigid transformation and residual deformation phases respectively.  During the compression process, we have adopted different precisions for the compression of positions depending on the dataset. Since the Gaussian positions in the N3DV dataset have a larger range, we have quantized the Gaussian positions of the N3DV dataset using uint32 precision and compressed all attributes with $\text{qp} = 10$. For the other two datasets, we have applied uint16 precision and $\text{qp} = 20$.  The H.264 encoder was configured using the x264 library with the following settings: I/P-frames only (no B-frames), 3 reference frames, color space in YUV4:4:4, and preset set to "medium."

Our experimental setup includes an Intel(R) Xeon(R) W-2245 CPU running at 3.90 GHz and an RTX 3090 graphics card. In the experiment, we conducted evaluations on a total of 12 sequences from the 4DGCPro and N3DV datasets. Also, we selected the Greeting and Umbrella sequences from the HiFi4G dataset. We selected the 48th view as the test view in the HiFi4G and 4DGCPro datasets, while the 0th view was chosen in the N3DV dataset. Due to the relatively high resolution of the HiFi4G and 4DGCPro datasets, we downsample them by a factor of 2 for training. In the experimental results, we reproduce and run the comparison methods on the HiFi4G and 4DGCPro datasets, and the results on the N3DV dataset are reported in the papers of 4DGC and HiCoM \cite{hicom2024}. Additionally, since the HPC \cite{zheng2024hpc} and $\text{V}^{3}$ methods cannot operate properly on datasets with backgrounds, we do not report their metrics on the N3DV dataset.  

\section{Additional Experimental Results}
\label{sec:appendixb}
\begin{figure*}[t]
\centering
\includegraphics[width=\linewidth]{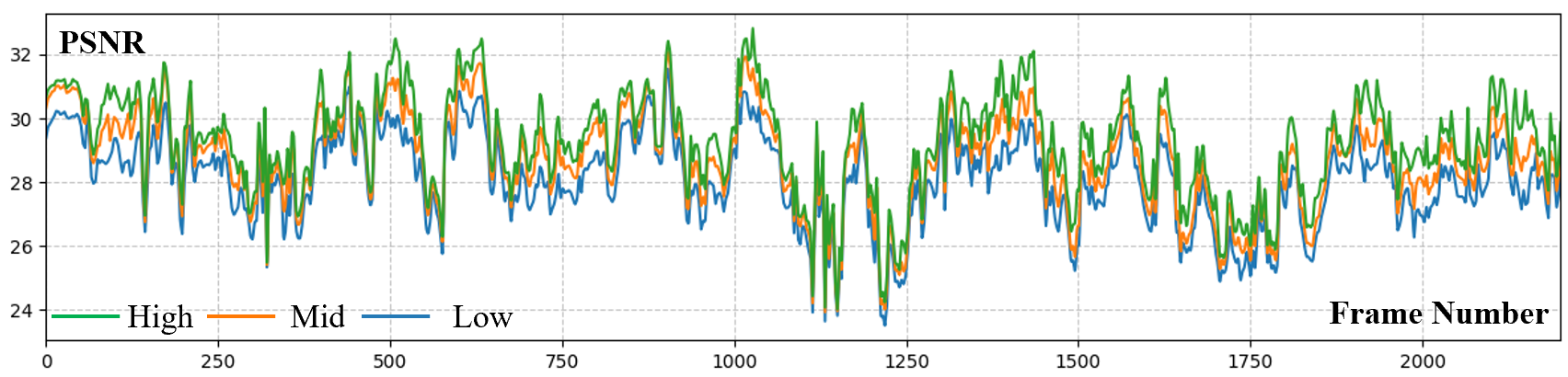}
\vspace{-1em}
\caption{The results of our method on long sequences. We show that the performance of our method does not decrease as the number of frames increases.}
\label{fig:long}
\end{figure*}
\begin{table*}[t]
\centering
\setlength{\tabcolsep}{6.9pt}
\caption{PSNR performance stability across three runs. Results are reported as the mean and standard deviation from three runs on the 4DGCPro dataset.}
\vspace{0.5em}
\label{tab:sup_t1}
\scalebox{0.85}{
\begin{tabular}{c|cccccc|c}
\hline
Method    & Dance1 & Dance2 & Coser1 & Coser2 & Boxing1 & Band1 & Mean       \\ \hline
Ours(High)      & 31.79 {\scriptsize$\pm$0.37}                                          & 29.78  {\scriptsize$\pm$0.11}                                             & 29.84 {\scriptsize$\pm$0.18}                                        & 25.65  {\scriptsize$\pm$0.24}                                           & 32.54  {\scriptsize$\pm$0.06}                                        & 27.19  {\scriptsize$\pm$0.15}                                         & 29.47 {\scriptsize$\pm$0.28} \\
Ours(Mid)      & 30.66  {\scriptsize$\pm$0.28}                                             & 29.05   {\scriptsize$\pm$0.14}                                           & 29.39  {\scriptsize$\pm$0.13}                                       & 25.40 {\scriptsize$\pm$0.21}                                             & 30.65  {\scriptsize$\pm$0.08}                                         & 26.93   {\scriptsize$\pm$0.10}                                       & 28.68 {\scriptsize$\pm$0.22}\\
Ours(Low)      & 29.26 {\scriptsize$\pm$0.17}                                            & 27.97  {\scriptsize$\pm$0.09}                                            & 28.60    {\scriptsize$\pm$0.08}                                   & 25.06  {\scriptsize$\pm$0.16}                                          & 28.99    {\scriptsize$\pm$0.05}                                        & 26.23    {\scriptsize$\pm$0.08}                                       & 27.69 {\scriptsize$\pm$0.14}\\ \hline
\end{tabular}
}
\end{table*}
\begin{table}[b]
\centering
\setlength{\tabcolsep}{9pt}
\caption{Runtime analysis on multiple platforms of rendering.}
\label{tab:sup_t5}
\scalebox{0.85}{
\begin{tabular}{c|ccc|ccc|ccc}
\hline
\multirow{2}{*}{Platform} & \multicolumn{3}{c|}{Desktop} & \multicolumn{3}{c|}{Tablet} & \multicolumn{3}{c}{Phone} \\ \cline{2-10} 
                          & High     & Mid     & Low     & High     & Mid     & Low    & High    & Mid    & Low    \\ \hline
Decoding(ms)              & 24       & 17      & 11      & 27       & 22      & 10     & 35      & 27     & 19     \\
Rendering(ms)             & 2.6      & 2.3     & 2.1     & 16       & 14      & 11     & 43      & 39     & 34     \\ \hline
\end{tabular}
}
\end{table}
\subsection{Addition Demonstrations of 4DGCPro}
\label{app:demonstrations}
\begin{figure*}[h]
\centering
\includegraphics[width=\linewidth]{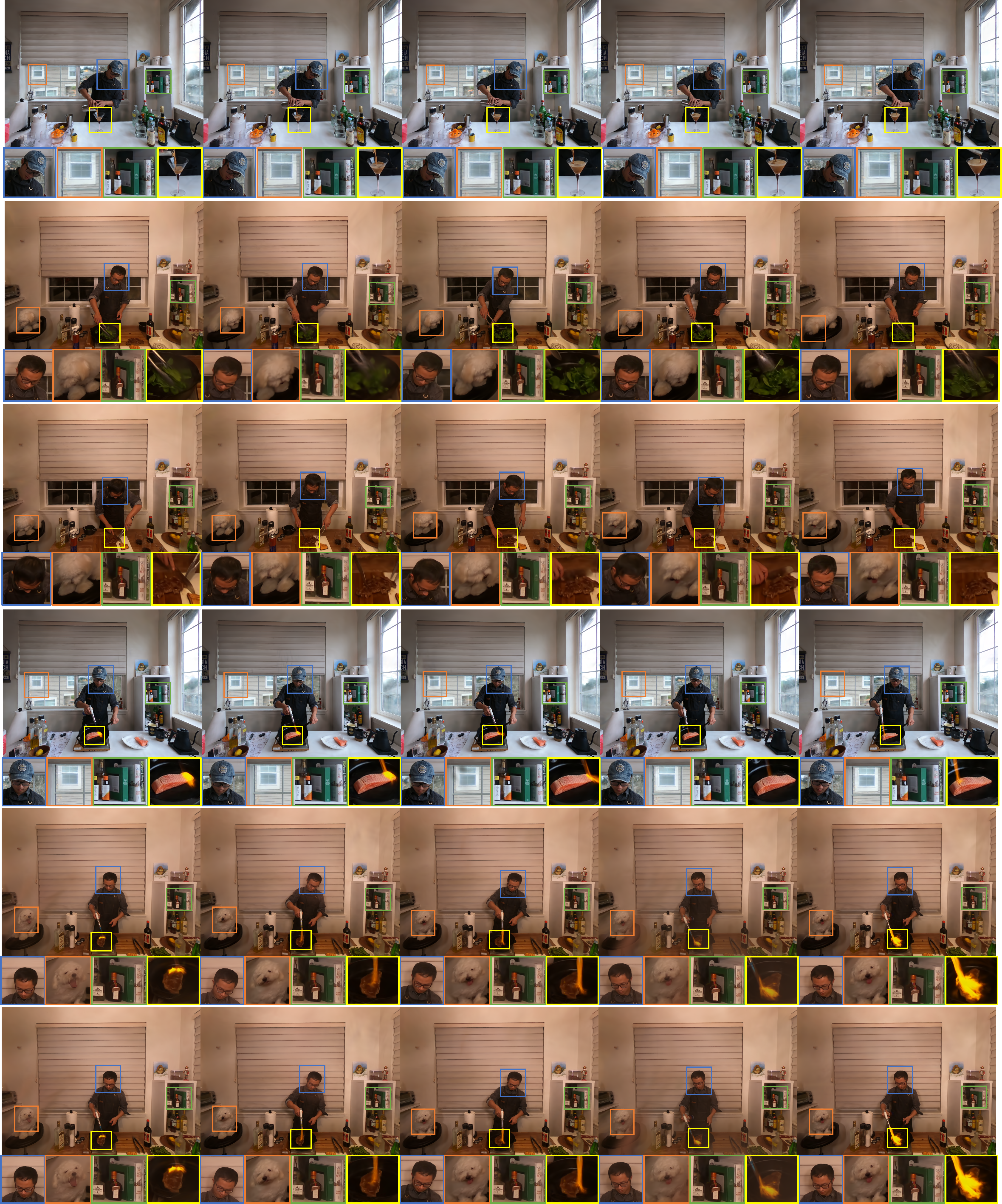}
\vspace{-1em}
\caption{Qualitative results of scenes from N3DV dataset. Frames shown are the $50^{\text{th}}$,
$100^{\text{th}}$, $150^{\text{th}}$, $200^{\text{th}}$, and $250^{\text{th}}$ from the test video.}
\label{fig:n3dv}
\end{figure*}

In this section, we additionally present some results of our method to comprehensively demonstrate the advantages of our method as much as possible. First of all, in Fig. \ref{fig:gallery}, the render gallery of our method is shown. It can be seen that our method can achieve high-quality reconstruction for scenarios with many complex motions. Fig. \ref{fig:hierarchical} shows the multi-bitrate results of our method under a single bitstream. From the figure, it can be observed that due to the high effectiveness of our hierarchical representation and layerwise supervision, 4DGCPro maintains excellent rendering quality at each level of the results. Fig. \ref{fig:long} presents the multi-bitrate results of our method in a long sequence. Our method can maintain a very high reconstruction quality in long sequences. As illustrated in Fig. \ref{fig:n3dv}, we present supplementary qualitative results for scenes from the N3DV \cite{li2022neural} dataset. These visual outcomes vividly showcase the resilience of our approach in accurately capturing and effectively representing dynamic scenes. 
In addition, to verify the stability of the method, we conducted multiple experiments under the same setting and reported the fluctuations of the results, as shown in Tab. \ref{tab:sup_t1}. 

\begin{table*}[t]
\centering
\setlength{\tabcolsep}{6.9pt}
\caption{Quantitative comparison of average PSNR values(dB) and model size(MB) across all sequences in the 4DGCPro dataset.}
\label{tab:sup_t2}
\scalebox{0.85}{
\begin{tabular}{c|cccccc|c}
\hline
Method    & Dance1 & Dance2 & Coser1 & Coser2 & Boxing1 & Band1 & Mean       \\ \hline
ReRF  & 30.01/0.98                                               & 26.30/2.17                                             & 27.30/1.32                                         & 24.35/1.99                                             & 31.12/1.46                                            &26.35/2.10                                           & 27.57/1.70 \\
HPC      & 30.78/0.83                                               & 26.95/0.92                                             & 26.90/0.86                                         & 24.01/1.29                                             & 31.00/1.15                                            & 26.45/1.42                                           & 27.68/1.08 \\
3DGStream   & 22.65/8.10                                               & 18.45/8.10                                          & 23.62/8.10                                      & 19.64/8.10                                             & 23.49/8.10                                           & 18.62/8.10                                           & 21.08/8.10\\
4DGC & 22.53/0.85                                              & 19.77/1.00                                             & 24.28/0.96                                         & 19.06/0.94                                             & 24.06/0.84                                            & 19.18/1.20                                           &21.48/0.97 \\
HiCoM      & 25.06/1.62                                               &23.81/1.72                                             & 24.55/2.07                                         & 21.51/3.87                                           & 28.87/2.92                                            & 24.07/3.47                                           & 24.65/2.61 \\ 
$\text{V}^{3}$      & 31.37/1.10                                               & 29.19/1.11                                              & 29.52/1.45                                        & 18.97/1.85                                             & \textbf{32.58}/1.61                                           & 27.11/2.46                                           & 28.11/1.60 \\ \hline
Ours(High)      & \textbf{31.79}/0.89                                              & \textbf{29.78}/0.78                                              & \textbf{29.84}/1.19                                         & \textbf{25.65}/1.80                                             & 32.54/1.33                                           & \textbf{27.19}/1.88                                           & \textbf{29.47}/1.31 \\
Ours(Mid)      & 30.66/0.45                                               & 29.05/0.39                                              & 29.39/0.60                                         & 25.40/0.91                                             & 30.65/0.66                                            & 26.93/0.94                                          & 28.68/0.66 \\
Ours(Low)      & 29.26/\textbf{0.22}                                               & 27.97/\textbf{0.21}                                              & 28.60/0.30                                       & 25.06/\textbf{0.45}                                             & 28.99/\textbf{0.33}                                            & 26.23/\textbf{0.47}                                           & 27.69/\textbf{0.33} \\ \hline
\end{tabular}
}
\end{table*}
\begin{table}[t]
\centering
\setlength{\tabcolsep}{6.9pt}
\caption{Quantitative comparison across two sequences in the HiFi4G dataset.}
\label{tab:sup_t3}
\scalebox{0.85}{
\begin{tabular}{c|cccc|cc}
\hline
\multirow{3}{*}{Method} & \multicolumn{2}{c}{Greeting}                                                                              & \multicolumn{2}{c|}{Umbrella}                                                                             & \multicolumn{2}{c}{Mean}                                                                                  \\ \cline{2-7} 
                        & \begin{tabular}[c]{@{}c@{}}PSNR\\ (dB)\end{tabular}$\uparrow$ & \begin{tabular}[c]{@{}c@{}}Size\\ (MB)\end{tabular}$\downarrow$ & \begin{tabular}[c]{@{}c@{}}PSNR\\ (dB)\end{tabular}$\uparrow$ & \begin{tabular}[c]{@{}c@{}}Size\\ (MB)\end{tabular}$\downarrow$ & \begin{tabular}[c]{@{}c@{}}PSNR\\ (dB)\end{tabular}$\uparrow$ & \begin{tabular}[c]{@{}c@{}}Size\\ (MB)\end{tabular}$\downarrow$ \\ \hline
ReRF                    & 29.90                                               & 0.96                                                & 30.69                                               & 0.98                                                & 30.30                                               & 0.97                                                \\
HPC                     & 35.47                                               & 0.75                                                & 32.81                                               & 0.69                                                & 34.14                                               & 0.72                                                \\
3DGStream               & 21.38                                               & 8.10                                                & 20.65                                               & 8.10                                                & 21.02                                               & 8.10                                                \\
4DGC                    & 20.94                                               & 0.99                                                & 21.15                                               & 0.89                                                & 21.05                                               & 0.94                                                \\
HiCoM                   & 29.12                                               & 1.70                                                & 29.61                                               & 2.18                                                & 29.37                                               & 1.94                                                \\
$\text{V}^{3}$                 & 37.06                                               & 0.89                                                & 35.45                                               & 0.95                                                & 36.26                                               & 0.92                                                \\ \hline
Ours(High)              & \textbf{37.21}                                               & 0.68                                                & \textbf{35.52}                                               & 0.81                                                & \textbf{36.38}                                               & 0.75                                                \\
Ours(Middle)            & 36.83                                               & 0.34                                                & 34.13                                               & 0.39                                                & 35.48                                               & 0.37                                                \\
Ours(Low)               & 35.96                                               & \textbf{0.17}                                                & 33.28                                               & \textbf{0.20}                                                & 34.62                                               & \textbf{0.19}                                                \\ \hline
\end{tabular}
}
\end{table}
We also conducted an efficiency analysis of our 4DGCPro across multiple platforms. The test platforms consist of an Ubuntu desktop PC featuring an Intel I9-10920X processor and an NVIDIA GeForce RTX 3090 GPU, an Apple iPad powered by an Apple M2 processor, and an Apple iPhone equipped with an A15 Bionic processor. As presented in Tab. \ref{tab:sup_t5}, we detail the time consumption of each thread within the rendering pipeline. During the decoding process, on the desktop, multi - threaded decoding combined with CUDA memory copying takes between \textbf{11ms} and \textbf{24ms}. For Apple's mobile devices, leveraging parallel decoding via compute shaders, these devices achieve a decoding time consumption comparable to that of the desktop (\textbf{22ms} on the tablet and \textbf{27ms} on the phone). Regarding the rendering thread, the desktop with a CUDA-enabled device demonstrates an extremely rapid rendering speed (\textbf{2.3ms}), while mobile devices are also capable of rendering at a satisfactory frame rate (\textbf{14ms} and \textbf{39ms}).
\subsection{Additional Comparison Results}
\label{sup:comparison}
We present the quantitative results for each scene from three datasets respectively in Tab. \ref{tab:sup_t2}, \ref{tab:sup_t3} and \ref{tab:sup_t4} to provide a more detailed comparison. We categorize the datasets into two groups: large-motion, background-free datasets such as 4DGCPro and HiFi4G \cite{hifi4g}, and small-motion, background-containing datasets like N3DV \cite{li2022neural}. Our method demonstrates a distinct advantage in the former group, which can be attributed to its precise motion modeling capabilities. In contrast, methods like 3DGStream \cite{sun20243dgstream} and 4DGC \cite{hu20254dgcrateaware4dgaussian} perform poorly. Specifically, they completely loses its modeling ability after processing only a very short sequence due to the error accumulation. This shortcoming stems from its inherent limitations of using only the first frame as a reference and being capable of modeling only rigid motion.
Notably, $\text{V}^{3}$ \cite{wang2024v} shows subpar performance on the coser2 sequence. The reason lies in the fact that NeuS2 \cite{neus2} fails to initialize the two interacting objects correctly during the training process, resulting in only a portion of the image being successfully modeled. Our approach addresses this issue by introducing residual NeuS2.
Within the N3DV dataset, given the small motion amplitudes, 3DGStream and 4DGC achieve good results. Nevertheless, our method still manages to deliver comparable performance.
\begin{table*}[t]
\centering
\setlength{\tabcolsep}{6.9pt}
\caption{Quantitative comparison of average PSNR values(dB) and model size(MB) across all sequences in the N3DV dataset.}
\label{tab:sup_t4}
\scalebox{0.85}{
\begin{tabular}{c|cccccc|c}
\hline
Method    & \begin{tabular}[c]{@{}c@{}}Coffee\\ Martini\end{tabular} & \begin{tabular}[c]{@{}c@{}}Cook\\ Spinach\end{tabular} & \begin{tabular}[c]{@{}c@{}}Cut\\ Beef\end{tabular} & \begin{tabular}[c]{@{}c@{}}Flame\\ Salmon\end{tabular} & \begin{tabular}[c]{@{}c@{}}Flame\\ Steak\end{tabular} & \begin{tabular}[c]{@{}c@{}}Sear\\ Steak\end{tabular} & Mean       \\ \hline
ReRF      & 26.24/0.79                                               & 31.23/0.84                                             & 31.82/0.81                                         & 26.80/0.78                                             & 32.08/0.91                                            & 30.03/0.51                                           & 29.71/0.77 \\
3DGStream & 27.96/8.00                                               & 32.88/8.05                                             & 32.99/8.19                                         & 28.52/8.07                                             & 33.41/8.19                                            & 33.58/8.16                                           & 31.54/8.11 \\
HiCoM  & 28.04/0.80                                               & 32.45/0.60                                             & 32.72/0.60                                         & 28.37/0.90                                             & 32.87/0.60                                            & 32.57/0.60                                           & 31.17/0.70 \\ 
4DGC  & \textbf{27.98}/0.58                                               & 32.81/0.44                                              & 33.03/0.47                                         & 28.49/0.51                                             & 33.58/0.44                                            & 33.60/0.50                                           & 31.58/0.49\\ \hline
Ours(High)      & 27.91/0.64                                              & \textbf{32.93}/0.67                                              & \textbf{33.10}/0.61                                         & \textbf{28.73}/0.61                                             & \textbf{33.77}/0.65                                           & \textbf{33.72}/0.66                                           & \textbf{31.64}/0.64 \\
Ours(Mid)      & 27.63/0.43                                               & 32.30/0.45                                              & 32.51/0.41                                         & 28.28/0.43                                             & 33.09/0.43                                            & 33.03/0.44                                          & 31.14/0.43 \\
Ours(Low)      & 26.88/\textbf{0.21}                                               & 31.95/\textbf{0.22}                                              & 32.17/\textbf{0.20}                                       & 27.95/\textbf{0.21}                                             & 32.49/\textbf{0.20}                                            & 32.64/\textbf{0.23}                                           & 30.68/\textbf{0.21} \\ \hline
\end{tabular}
}
\end{table*}
\subsection{Additional Ablation Studies}
\label{sec:sup_ablation}
In this section, we conducted additional experiments to validate the efficacy of motion decomposition and layer-wise supervision, as shown in Tab. \ref{tab:sup_t6}. The results reveal that motion decomposition enables precise modeling of dynamic scenes by distinctly separating the processes of rigid transformation and residual deformation. However, simultaneously training these two components makes the positions of Gaussians less accurate and increases the training difficulty, and accurate results cannot be obtained with the same number of training steps. 
Regarding layer-wise supervision, while it has a marginal impact on the reconstruction quality of complete Gaussians, it yields a substantial boost in the quality of lower-level Gaussians.
\begin{table}[t]
\centering
\setlength{\tabcolsep}{6.9pt}
\caption{Results of more ablation studies.}
\label{tab:sup_t6}
\scalebox{0.85}{
\begin{tabular}{c|cc|cc|cc}
\hline
\multirow{2}{*}{Ablation Studies} & \multicolumn{2}{c|}{High} & \multicolumn{2}{c|}{Mid} & \multicolumn{2}{c}{Low} \\ \cline{2-7} 
                                                                            & PSNR(dB)    & Size(MB)    & PSNR(dB)    & Size(MB)   & PSNR(dB)   & Size(MB)   \\ \hline
w/o Motion Decomposition                                                           & 28.84       & 1.31        & 28.17       & 0.66       & 27.04      & 0.33       \\
w/o Layer-wise Supervision                                                  & 29.53       & 1.33        & 26.49       & 0.67       & 24.98      & 0.34       \\
Ours Full                                                                   & \textbf{29.47}       & 1.31        & \textbf{28.68}       & 0.66       & \textbf{27.69}      & 0.33       \\ \hline
\end{tabular}
}
\end{table}


\clearpage
\section*{NeurIPS Paper Checklist}



\begin{enumerate}

\item {\bf Claims}
    \item[] Question: Do the main claims made in the abstract and introduction accurately reflect the paper's contributions and scope?
    \item[] Answer: \answerYes{} 
    \item[] Justification: The main claims made in the abstract and introduction accurately reflect the
 paper’s contributions and scope.
    \item[] Guidelines:
    \begin{itemize}
        \item The answer NA means that the abstract and introduction do not include the claims made in the paper.
        \item The abstract and/or introduction should clearly state the claims made, including the contributions made in the paper and important assumptions and limitations. A No or NA answer to this question will not be perceived well by the reviewers. 
        \item The claims made should match theoretical and experimental results, and reflect how much the results can be expected to generalize to other settings. 
        \item It is fine to include aspirational goals as motivation as long as it is clear that these goals are not attained by the paper. 
    \end{itemize}

\item {\bf Limitations}
    \item[] Question: Does the paper discuss the limitations of the work performed by the authors?
    \item[] Answer: \answerYes{} 
    \item[] Justification: The limitations has been discussed in the Discussion section.
    \item[] Guidelines:
    \begin{itemize}
        \item The answer NA means that the paper has no limitation while the answer No means that the paper has limitations, but those are not discussed in the paper. 
        \item The authors are encouraged to create a separate "Limitations" section in their paper.
        \item The paper should point out any strong assumptions and how robust the results are to violations of these assumptions (e.g., independence assumptions, noiseless settings, model well-specification, asymptotic approximations only holding locally). The authors should reflect on how these assumptions might be violated in practice and what the implications would be.
        \item The authors should reflect on the scope of the claims made, e.g., if the approach was only tested on a few datasets or with a few runs. In general, empirical results often depend on implicit assumptions, which should be articulated.
        \item The authors should reflect on the factors that influence the performance of the approach. For example, a facial recognition algorithm may perform poorly when image resolution is low or images are taken in low lighting. Or a speech-to-text system might not be used reliably to provide closed captions for online lectures because it fails to handle technical jargon.
        \item The authors should discuss the computational efficiency of the proposed algorithms and how they scale with dataset size.
        \item If applicable, the authors should discuss possible limitations of their approach to address problems of privacy and fairness.
        \item While the authors might fear that complete honesty about limitations might be used by reviewers as grounds for rejection, a worse outcome might be that reviewers discover limitations that aren't acknowledged in the paper. The authors should use their best judgment and recognize that individual actions in favor of transparency play an important role in developing norms that preserve the integrity of the community. Reviewers will be specifically instructed to not penalize honesty concerning limitations.
    \end{itemize}

\item {\bf Theory assumptions and proofs}
    \item[] Question: For each theoretical result, does the paper provide the full set of assumptions and a complete (and correct) proof?
    \item[] Answer: \answerYes{} 
    \item[] Justification: All the theorems, formulas in the paper are numbered and cross-referenced.
    \item[] Guidelines:
    \begin{itemize}
        \item The answer NA means that the paper does not include theoretical results. 
        \item All the theorems, formulas, and proofs in the paper should be numbered and cross-referenced.
        \item All assumptions should be clearly stated or referenced in the statement of any theorems.
        \item The proofs can either appear in the main paper or the supplemental material, but if they appear in the supplemental material, the authors are encouraged to provide a short proof sketch to provide intuition. 
        \item Inversely, any informal proof provided in the core of the paper should be complemented by formal proofs provided in appendix or supplemental material.
        \item Theorems and Lemmas that the proof relies upon should be properly referenced. 
    \end{itemize}

    \item {\bf Experimental result reproducibility}
    \item[] Question: Does the paper fully disclose all the information needed to reproduce the main experimental results of the paper to the extent that it affects the main claims and/or conclusions of the paper (regardless of whether the code and data are provided or not)?
    \item[] Answer: \answerYes{} 
    \item[] Justification: : We will open source our code to replicate our results.
    \item[] Guidelines:
    \begin{itemize}
        \item The answer NA means that the paper does not include experiments.
        \item If the paper includes experiments, a No answer to this question will not be perceived well by the reviewers: Making the paper reproducible is important, regardless of whether the code and data are provided or not.
        \item If the contribution is a dataset and/or model, the authors should describe the steps taken to make their results reproducible or verifiable. 
        \item Depending on the contribution, reproducibility can be accomplished in various ways. For example, if the contribution is a novel architecture, describing the architecture fully might suffice, or if the contribution is a specific model and empirical evaluation, it may be necessary to either make it possible for others to replicate the model with the same dataset, or provide access to the model. In general. releasing code and data is often one good way to accomplish this, but reproducibility can also be provided via detailed instructions for how to replicate the results, access to a hosted model (e.g., in the case of a large language model), releasing of a model checkpoint, or other means that are appropriate to the research performed.
        \item While NeurIPS does not require releasing code, the conference does require all submissions to provide some reasonable avenue for reproducibility, which may depend on the nature of the contribution. For example
        \begin{enumerate}
            \item If the contribution is primarily a new algorithm, the paper should make it clear how to reproduce that algorithm.
            \item If the contribution is primarily a new model architecture, the paper should describe the architecture clearly and fully.
            \item If the contribution is a new model (e.g., a large language model), then there should either be a way to access this model for reproducing the results or a way to reproduce the model (e.g., with an open-source dataset or instructions for how to construct the dataset).
            \item We recognize that reproducibility may be tricky in some cases, in which case authors are welcome to describe the particular way they provide for reproducibility. In the case of closed-source models, it may be that access to the model is limited in some way (e.g., to registered users), but it should be possible for other researchers to have some path to reproducing or verifying the results.
        \end{enumerate}
    \end{itemize}

\item {\bf Open access to data and code}
    \item[] Question: Does the paper provide open access to the data and code, with sufficient instructions to faithfully reproduce the main experimental results, as described in supplemental material?
    \item[] Answer: \answerYes{} 
    \item[] Justification: The paper provides open access to both the data and the code.
    \item[] Guidelines:
    \begin{itemize}
        \item The answer NA means that paper does not include experiments requiring code.
        \item Please see the NeurIPS code and data submission guidelines (\url{https://nips.cc/public/guides/CodeSubmissionPolicy}) for more details.
        \item While we encourage the release of code and data, we understand that this might not be possible, so “No” is an acceptable answer. Papers cannot be rejected simply for not including code, unless this is central to the contribution (e.g., for a new open-source benchmark).
        \item The instructions should contain the exact command and environment needed to run to reproduce the results. See the NeurIPS code and data submission guidelines (\url{https://nips.cc/public/guides/CodeSubmissionPolicy}) for more details.
        \item The authors should provide instructions on data access and preparation, including how to access the raw data, preprocessed data, intermediate data, and generated data, etc.
        \item The authors should provide scripts to reproduce all experimental results for the new proposed method and baselines. If only a subset of experiments are reproducible, they should state which ones are omitted from the script and why.
        \item At submission time, to preserve anonymity, the authors should release anonymized versions (if applicable).
        \item Providing as much information as possible in supplemental material (appended to the paper) is recommended, but including URLs to data and code is permitted.
    \end{itemize}

\item {\bf Experimental setting/details}
    \item[] Question: Does the paper specify all the training and test details (e.g., data splits, hyperparameters, how they were chosen, type of optimizer, etc.) necessary to understand the results?
    \item[] Answer: \answerYes{} 
    \item[] Justification: The details of training and testing are provided in the Sec. \ref{sec:appendixa}. 
    \item[] Guidelines:
    \begin{itemize}
        \item The answer NA means that the paper does not include experiments.
        \item The experimental setting should be presented in the core of the paper to a level of detail that is necessary to appreciate the results and make sense of them.
        \item The full details can be provided either with the code, in appendix, or as supplemental material.
    \end{itemize}

\item {\bf Experiment statistical significance}
    \item[] Question: Does the paper report error bars suitably and correctly defined or other appropriate information about the statistical significance of the experiments?
    \item[] Answer: \answerNo{} 
    \item[] Justification: We ran all our experiments 3 times and reported the mean metrics. However, since some of the results are directly cited from other papers, we did not include error bars in the main text to maintain consistency. Nevertheless, we reported the standard deviation of the available results in the appendix, please see Tab. \ref{tab:sup_t1}.
    \item[] Guidelines:
    \begin{itemize}
        \item The answer NA means that the paper does not include experiments.
        \item The authors should answer "Yes" if the results are accompanied by error bars, confidence intervals, or statistical significance tests, at least for the experiments that support the main claims of the paper.
        \item The factors of variability that the error bars are capturing should be clearly stated (for example, train/test split, initialization, random drawing of some parameter, or overall run with given experimental conditions).
        \item The method for calculating the error bars should be explained (closed form formula, call to a library function, bootstrap, etc.)
        \item The assumptions made should be given (e.g., Normally distributed errors).
        \item It should be clear whether the error bar is the standard deviation or the standard error of the mean.
        \item It is OK to report 1-sigma error bars, but one should state it. The authors should preferably report a 2-sigma error bar than state that they have a 96\% CI, if the hypothesis of Normality of errors is not verified.
        \item For asymmetric distributions, the authors should be careful not to show in tables or figures symmetric error bars that would yield results that are out of range (e.g. negative error rates).
        \item If error bars are reported in tables or plots, The authors should explain in the text how they were calculated and reference the corresponding figures or tables in the text.
    \end{itemize}

\item {\bf Experiments compute resources}
    \item[] Question: For each experiment, does the paper provide sufficient information on the computer resources (type of compute workers, memory, time of execution) needed to reproduce the experiments?
    \item[] Answer: \answerYes{} 
    \item[] Justification: The details of compute resources are provided in Sec. \ref{sec:appendixa}.
    \item[] Guidelines:
    \begin{itemize}
        \item The answer NA means that the paper does not include experiments.
        \item The paper should indicate the type of compute workers CPU or GPU, internal cluster, or cloud provider, including relevant memory and storage.
        \item The paper should provide the amount of compute required for each of the individual experimental runs as well as estimate the total compute. 
        \item The paper should disclose whether the full research project required more compute than the experiments reported in the paper (e.g., preliminary or failed experiments that didn't make it into the paper). 
    \end{itemize}
    
\item {\bf Code of ethics}
    \item[] Question: Does the research conducted in the paper conform, in every respect, with the NeurIPS Code of Ethics \url{https://neurips.cc/public/EthicsGuidelines}?
    \item[] Answer: \answerYes{} 
    \item[] Justification: The research conducted in the paper conform, in every respect, with the
NeurIPS Code of Ethics.
    \item[] Guidelines:
    \begin{itemize}
        \item The answer NA means that the authors have not reviewed the NeurIPS Code of Ethics.
        \item If the authors answer No, they should explain the special circumstances that require a deviation from the Code of Ethics.
        \item The authors should make sure to preserve anonymity (e.g., if there is a special consideration due to laws or regulations in their jurisdiction).
    \end{itemize}

\item {\bf Broader impacts}
    \item[] Question: Does the paper discuss both potential positive societal impacts and negative societal impacts of the work performed?
    \item[] Answer: \answerNo{} 
    \item[] Justification: This study mainly focuses on the innovation of volumetric video reconstruction and compression algorithms, aiming to improve the reconstruction accuracy and efficiency and expand the generality of the algorithms. The datasets used in the research are all publicly available standard datasets and self-made datasets that will be made public, which do not contain any data involving personal privacy or sensitive information. 
    \item[] Guidelines:
    \begin{itemize}
        \item The answer NA means that there is no societal impact of the work performed.
        \item If the authors answer NA or No, they should explain why their work has no societal impact or why the paper does not address societal impact.
        \item Examples of negative societal impacts include potential malicious or unintended uses (e.g., disinformation, generating fake profiles, surveillance), fairness considerations (e.g., deployment of technologies that could make decisions that unfairly impact specific groups), privacy considerations, and security considerations.
        \item The conference expects that many papers will be foundational research and not tied to particular applications, let alone deployments. However, if there is a direct path to any negative applications, the authors should point it out. For example, it is legitimate to point out that an improvement in the quality of generative models could be used to generate deepfakes for disinformation. On the other hand, it is not needed to point out that a generic algorithm for optimizing neural networks could enable people to train models that generate Deepfakes faster.
        \item The authors should consider possible harms that could arise when the technology is being used as intended and functioning correctly, harms that could arise when the technology is being used as intended but gives incorrect results, and harms following from (intentional or unintentional) misuse of the technology.
        \item If there are negative societal impacts, the authors could also discuss possible mitigation strategies (e.g., gated release of models, providing defenses in addition to attacks, mechanisms for monitoring misuse, mechanisms to monitor how a system learns from feedback over time, improving the efficiency and accessibility of ML).
    \end{itemize}
    
\item {\bf Safeguards}
    \item[] Question: Does the paper describe safeguards that have been put in place for responsible release of data or models that have a high risk for misuse (e.g., pretrained language models, image generators, or scraped datasets)?
    \item[] Answer: \answerNo{} 
    \item[] Justification: The volumetric video reconstruction and compression algorithms proposed in this study, along with the related data, carry a low risk of being maliciously misused. Both the publicly available standard datasets used in the study and the self-made datasets planned for public release do not involve privacy-sensitive information or data content that could be directly used for harmful purposes. Additionally, the algorithms of this study mainly focus on improving the reconstruction accuracy and compression efficiency, and do not have the direct ability to generate misleading content or infringe on personal rights. At the current stage, no high-risk scenarios requiring special security safeguards have been found, so no specific security protection mechanisms have been designed for the release of data or models. However, if potential risks are discovered in the future, we will actively explore and implement appropriate protective measures. 
    \item[] Guidelines:
    \begin{itemize}
        \item The answer NA means that the paper poses no such risks.
        \item Released models that have a high risk for misuse or dual-use should be released with necessary safeguards to allow for controlled use of the model, for example by requiring that users adhere to usage guidelines or restrictions to access the model or implementing safety filters. 
        \item Datasets that have been scraped from the Internet could pose safety risks. The authors should describe how they avoided releasing unsafe images.
        \item We recognize that providing effective safeguards is challenging, and many papers do not require this, but we encourage authors to take this into account and make a best faith effort.
    \end{itemize}

\item {\bf Licenses for existing assets}
    \item[] Question: Are the creators or original owners of assets (e.g., code, data, models), used in the paper, properly credited and are the license and terms of use explicitly mentioned and properly respected?
    \item[] Answer: \answerYes{} 
    \item[] Justification: The creators or original owners of assets (e.g., code, data, models), used in the paper, are properly credited, and the license and terms of use explicitly are mentioned and properly respected.
    \item[] Guidelines:
    \begin{itemize}
        \item The answer NA means that the paper does not use existing assets.
        \item The authors should cite the original paper that produced the code package or dataset.
        \item The authors should state which version of the asset is used and, if possible, include a URL.
        \item The name of the license (e.g., CC-BY 4.0) should be included for each asset.
        \item For scraped data from a particular source (e.g., website), the copyright and terms of service of that source should be provided.
        \item If assets are released, the license, copyright information, and terms of use in the package should be provided. For popular datasets, \url{paperswithcode.com/datasets} has curated licenses for some datasets. Their licensing guide can help determine the license of a dataset.
        \item For existing datasets that are re-packaged, both the original license and the license of the derived asset (if it has changed) should be provided.
        \item If this information is not available online, the authors are encouraged to reach out to the asset's creators.
    \end{itemize}

\item {\bf New assets}
    \item[] Question: Are new assets introduced in the paper well documented and is the documentation provided alongside the assets?
    \item[] Answer: \answerNA{} 
    \item[] Justification: N/A.
    \item[] Guidelines:
    \begin{itemize}
        \item The answer NA means that the paper does not release new assets.
        \item Researchers should communicate the details of the dataset/code/model as part of their submissions via structured templates. This includes details about training, license, limitations, etc. 
        \item The paper should discuss whether and how consent was obtained from people whose asset is used.
        \item At submission time, remember to anonymize your assets (if applicable). You can either create an anonymized URL or include an anonymized zip file.
    \end{itemize}

\item {\bf Crowdsourcing and research with human subjects}
    \item[] Question: For crowdsourcing experiments and research with human subjects, does the paper include the full text of instructions given to participants and screenshots, if applicable, as well as details about compensation (if any)? 
    \item[] Answer: \answerNA{} 
    \item[] Justification: N/A.
    \item[] Guidelines:
    \begin{itemize}
        \item The answer NA means that the paper does not involve crowdsourcing nor research with human subjects.
        \item Including this information in the supplemental material is fine, but if the main contribution of the paper involves human subjects, then as much detail as possible should be included in the main paper. 
        \item According to the NeurIPS Code of Ethics, workers involved in data collection, curation, or other labor should be paid at least the minimum wage in the country of the data collector. 
    \end{itemize}

\item {\bf Institutional review board (IRB) approvals or equivalent for research with human subjects}
    \item[] Question: Does the paper describe potential risks incurred by study participants, whether such risks were disclosed to the subjects, and whether Institutional Review Board (IRB) approvals (or an equivalent approval/review based on the requirements of your country or institution) were obtained?
    \item[] Answer: \answerNA{} 
    \item[] Justification: N/A.
    \item[] Guidelines:
    \begin{itemize}
        \item The answer NA means that the paper does not involve crowdsourcing nor research with human subjects.
        \item Depending on the country in which research is conducted, IRB approval (or equivalent) may be required for any human subjects research. If you obtained IRB approval, you should clearly state this in the paper. 
        \item We recognize that the procedures for this may vary significantly between institutions and locations, and we expect authors to adhere to the NeurIPS Code of Ethics and the guidelines for their institution. 
        \item For initial submissions, do not include any information that would break anonymity (if applicable), such as the institution conducting the review.
    \end{itemize}

\item {\bf Declaration of LLM usage}
    \item[] Question: Does the paper describe the usage of LLMs if it is an important, original, or non-standard component of the core methods in this research? Note that if the LLM is used only for writing, editing, or formatting purposes and does not impact the core methodology, scientific rigorousness, or originality of the research, declaration is not required.
    \item[] Answer: \answerNA{} 
    \item[] Justification: N/A.
    \item[] Guidelines:
    \begin{itemize}
        \item The answer NA means that the core method development in this research does not involve LLMs as any important, original, or non-standard components.
        \item Please refer to our LLM policy (\url{https://neurips.cc/Conferences/2025/LLM}) for what should or should not be described.
    \end{itemize}

\end{enumerate}

\end{document}